\documentclass[review,authoryear,12pt]{elsarticle}
\usepackage{lineno}
\modulolinenumbers[5]
\label{sec:packages}
\usepackage{booktabs}  % Me addf
\usepackage{multirow}
\usepackage{bm}
\usepackage{array}
\usepackage{graphicx}
\usepackage{subfigure}
\usepackage{amssymb}
\usepackage{amsmath}
\usepackage[pagebackref=false,breaklinks=true,letterpaper=true,bookmarks=false]{hyperref}
\usepackage{verbatim}
\usepackage{mathrsfs}
\usepackage{setspace}
\usepackage{amsmath}
\usepackage{epsfig}
\usepackage{amsfonts}
\usepackage{verbatim}
\usepackage{amssymb}
\usepackage{amsthm}
\usepackage{rotating}
\usepackage{graphicx}
\usepackage{subfigure}
\usepackage{booktabs}
\usepackage[linesnumbered,ruled,lined]{algorithm2e}
\usepackage{array}

\newcommand{\eat}[1]{}

\usepackage{bm}
\usepackage{multirow}
\usepackage{rotating}
\usepackage{color}
\usepackage[noend]{algpseudocode}
\usepackage{verbatim}
\usepackage{mdwtab}
\usepackage{color}
\usepackage{marginnote}
\usepackage{url}

\graphicspath{{Figures/}}

\label{definition}

\def\0{{\bf 0}}
\def\1{{\bf 1}}

\def\etal{{\em et al.}}
\def\eg{{\em e.g.}}
\def\ie{{\em i.e.}}

\def\etal{{\em et al.\/}\,}

\makeatletter

\if false
\def\cortext[#1]#2{\g@addto@macro\@cornotes{%
		\refstepcounter{cnote}\elsLabel{#1}%
		\def\thefootnote{\ifcase\thecnote\or$\dagger$\or
			$\ast$\fi}%
		\footnotetext{#2}}}

\def\corref#1{\edef\cnotenum{\elsRef{#1}}%
	\edef\@corref{\ifcase\cnotenum\or
		$\dagger$\or$\ast$\fi\hskip-1pt}}
\fi

\makeatother

\journal{Medical Image Analysis}

\begin{document}

\begin{frontmatter}

\title{\textbf{Towards Evaluating the Robustness of Deep Diagnostic Models by Adversarial Attack}}

\author[nuaa]{Mengting~Xu}\ead{xumengting@nuaa.edu.cn}
\author[nuaa]{Tao~Zhang}\ead{lrhselous@nuaa.edu.cn}
\author[nuaa]{Zhongnian~Li}\ead{zhongnianli@163.com}
\author[unc]{Mingxia~Liu\corref{cor}}\ead{mxliu1226@gmail.com}
\author[nuaa]{Daoqiang~Zhang\corref{cor}}\ead{dqzhang@nuaa.edu.cn}
%\author[unc]{Jun~Zhang\corref{equal}}\ead{xdzhangjun@gmail.com}
%\author[unc]{Ehsan Adeli}\ead{eadeli@gmail.com}
%\author[unc,korea]{Dinggang Shen\corref{cor}} \ead{dgshen@med.unc.edu}
\cortext[cor]{Corresponding authors: Mingxia Liu and Daoqiang Zhang.}
%\cortext[equal]{These authors contribute equally to this study}
\address[nuaa]{College of Computer Science and Technology, Nanjing University of Aeronautics and Astronautics, Nanjing 211106, China}
\address[unc]{Department of Radiology and BRIC, University of North Carolina at Chapel Hill, North Carolina 27599, USA}
%\address[korea]{Department of Brain and Cognitive Engineering, Korea University, Seoul 02841, Republic of Korea}

\begin{abstract}
Deep learning models (with neural networks) have been widely used in challenging tasks such as computer-aided disease diagnosis based on medical images. 
Recent studies have shown deep diagnostic models may not be robust in the inference process and may pose severe security concerns in clinical practice. Among all the factors that make the model not robust, the most serious one is adversarial examples. The so-called ``adversarial example'' is a well-designed perturbation that is not easily perceived by humans but results in a false output of deep diagnostic models with high confidence. 
In this paper, we evaluate the robustness of deep diagnostic models by adversarial attack.
%we are concerned about whether the representative deep diagnostic models are still reliable under adversarial attacks.
Specifically, we have performed two types of adversarial attacks to three deep diagnostic models in both single-label and multi-label classification tasks, and found that these models are not reliable when attacked by adversarial example. We have further explored how adversarial examples attack the models, by analyzing their quantitative classification results, intermediate features, discriminability of features and correlation of estimated labels for both original/clean images and those adversarial ones. 
We have also designed two new defense methods to handle adversarial examples in deep diagnostic models, \ie, Multi-Perturbations Adversarial Training (MPAdvT) and Misclassification-Aware Adversarial Training (MAAdvT). The experimental results have shown that the use of defense methods can significantly improve the robustness of deep diagnostic models against adversarial attacks.
\end{abstract}


\end{frontmatter}

%\linenumbers
%% %% %% ---------------------------------------------------%% %% ^_^
\section{Introduction}
\label{S1}
Deep learning algorithms, powered by advances in neural network structures and large amounts of data, have shown high performance (even exceeding human potential) in healthcare applications.
There are many impressive examples of deep learning with excellent performances in medical tasks of radiology~\citep{gale2017detecting,rajpurkar2017chexnet}, pathology~\citep{bejnordi2017diagnostic}, dermatology~\citep{esteva2017dermatologist} and ophthalmology~\citep{gulshan2016development}. Especially, deep learning models have attained state-of-the-art performance in many challenges on medical image analysis, such as segmentation of lesions in the brain  (top ranked in BRATS and ISLES)~\citep{ghafoorian2016non}, prostate segmentation (top ranked in PROMISE challenge)~\citep{zhou2018fine}, and disease diagnosis~\citep{louis20162016,liu2018joint,liu2018landmark,lian2018hierarchical,jie2020designing,wang2019spatial,zhang2020survey}. 
Besides, the U.S. Food and Drug Administration (FDA) has approved diagnostic procedures using artificial intelligence (AI) technologies to detect greater than moderate levels of diabetic retinopathy without requiring doctors' assistance~\citep{finlayson2018adversarial}.

However, a key problem that cannot be ignored in deep diagnostic models is its robustness and reliability. Deep models often produce incomprehensible mistakes under noisy environments, thus leading to unexpected serious consequences. 
Zheng~\etal~\citep{zheng2016improving} have illustrated that current feature embeddings and class labels are not robust to a large class of small perturbations. Recent studies have also shown that deep models are highly vulnerable to adversarial examples, \ie, slightly perturbed images resembling original images but maliciously designed to fool pre-trained models~\citep{goodfellow2014explaining,szegedy2013intriguing,dong2018boosting,moosavi2017universal,poursaeed2018generative,finlayson2019adversarial}.

Medical safety is paramount in clinical practice, and therefore the vulnerabilities of deep models and the security threats they pose from deploying these algorithms in virtual and physical environments have attracted widespread attention.
If the doctor is not involved in the diagnosis process at all (which now has legal sanction in at least one setting via FDA, with many more to likely follow), we are forced to consider the question of how unreliable deep diagnostic models are attacked by adversarial perturbations, as this problem may lead to new opportunities for fraud and harm.
For example, diagnostic errors will make the disease worse for patients and harm the reputation of healthcare departments.
Even with a human in the loop, any clinical system that leverages a machine learning algorithm for diagnosis, decision-making, or reimbursement could be manipulated with adversarial examples~\citep{finlayson2018adversarial}.

\textit{\textbf{Will deep diagnostic models still be reliable under adversarial attack? What is the performance of these models when confronted with adversarial perturbations? Whether the robustness of deep diagnostic models can be improved?}}

%==> our works
To explore these questions, we evaluate the robustness of three representative deep diagnostic models with medical images, including (1) IPMI2019-AttnMel for Melanoma~\citep{yan2019melanoma,esteva2017dermatologist}, (2) Inception\_v3 for Diabetic Retinopathy (with a dataset called Messidor)~\citep{gulshan2016development,sahlsten2019deep}, and (3) CheXNet for 14 diseases on ChestX-Ray~\citep{wang2017chestx,huang2017densely,baltruschat2019comparison,pasa2019efficient}, by extending previous results on adversarial examples. 
In the experiments, we evaluate the robustness of these three models with adversarial attacks from four aspects. We first record the performance of these models by analyzing the decrease of diagnostic accuracy (ACC)\footnote{Accuracy indicates the percentage of images on which a trained model outputs its true label.}, as well as the increase in fooling ratio (FR)\footnote{Fooling ratio indicates the percentage of images on which a trained model changes its prediction label after the images are perturbed.}. We further visualize the feature maps generated by each model before and after adversarial attacks, and also explore the effectiveness of adversarial perturbations on outputs of different network layers. We further study the relationship between adversarial and original labels. These results indicate that these representative deep diagnostic models are vulnerable to adversarial perturbations in three tasks of binary, multi-class and multi-label classification. 
This encourages us to think carefully before deploying deep diagnostic models to the clinical systems and urges us to explore more robust medical models.
Besides, we create a new dataset (called \textit{Robust-Benchmark}) to evaluate the robustness of deep diagnostic models against common perturbations comprehensively.
Considering the vulnerability of deep diagnostic models attacked by adversarial examples, we have designed two new defense methods to handle this problem in deep diagnostic models, \ie, {\textbf{Multi-Perturbations Adversarial Training (MPAdvT)}} and {\textbf{Misclassification-Aware Adversarial Training (MAAdvT)}.  
We compared our proposed MPAdvT and MAAdvT with the conventional  adversarial training~\citep{madry2017towards}, and the results indicate that our methods can significantly improve the robustness of deep diagnostic models.
}

\textbf{The main contributions of this work are summarized as follows:}
\begin{itemize}
\item We evaluated the robustness of three representative deep diagnostic models in three tasks of binary, multi-class and multi-label classification (\ie, IPMI2019-AttnMel for melanoma classification~\citep{yan2019melanoma}, Inception\_v3 for detection of diabetic retinopathy~\citep{gulshan2016development}, and CheXNet for classification of 14 types of diseases on ChestX-Ray~\citep{huang2017densely}). To this end, we performed comprehensive analysis from four perspectives: 1) quantitative classification results, 2) intermediate features, 3) discriminability of features, and 4) correlation of estimated labels.
\item In addition to evaluating the robustness of deep diagnostic models against adversarial attacks, we further evaluated the robustness of models against common perturbations by creating a new dataset (called \textit{Robust-Benchmark}) of medical images. 
This dataset can be used as a general dataset to evaluate the robustness of deep diagnostic models in a standard way.
\item We proposed two new defense methods to handle adversarial examples in deep diagnostic models, called Multi-Perturbations Adversarial Training (MPAdvT) and Misclassification-Aware Adversarial Training (MAAdvT), respectively. 
Experimental results have shown that the proposed defense methods can effectively improve the robustness of deep diagnostic models. 
\end{itemize}

%% %% %% ---------------------------------------------------%% %% ^_^
\section{Related Work}
\label{S2}

In this section, we first briefly introduce recent development of deep learning in the field of medical image analysis and prior work on disease diagnosis. We then review recent adversarial attack and defense methods on nature and medical images.

\subsection{Deep Diagnostic Models for Medical Image Analysis}
%==>first introduce deep learning in medical images
Deep diagnostic classification frameworks have emerged for disease diagnosis over the past few years.
Now we would like to introduce three successful applications of deep learning models in medical image analysis.

Melanoma is one of the deadliest skin cancers in the world. However, accurate diagnosis of melanoma is non-trivial and requires expert human knowledge. Many automatic algorithms were proposed to classify melanoma from dermoscopy images~\citep{yan2019melanoma}. Particularly, deep learning methods have been used in top-performing approaches~\citep{gutman2016skin,codella2018skin}.
A challenge at the International Symposium on Biomedical Imaging (ISBI) 2016, hosted by the International Skin Imaging Collaboration (ISIC)  is completed with 79 submissions from a group of 38 participants, making this the largest standardized and comparative study for melanoma diagnosis in dermoscopic images to date~\citep{gutman2016skin}.
Esteva \etal~\cite{esteva2017dermatologist} collected a large dataset for this challenging task to improve the generalization capability of medical practitioners and utilize an inception\_v3 architecture for disease diagnosis with the ambition for low-cost universal access to vital diagnostic care. 
Yan \etal~\citep{yan2019melanoma} proposed an attention-based method for melanoma recognition, which is the first to introduce an end-to-end trainable attention module with regularization for melanoma recognition.

Diabetic retinopathy (DR) is also a common disease, which is the leading cause of blindness in the working-age population of the developed world. Automated grading of diabetic retinopathy has potential benefits such as increased efficiency, reproducibility, and coverage of screening programs, reducing barriers to access, and improving patient outcomes by providing early detection and treatment. To maximize the clinical utility of automated grading, an algorithm to detect referable diabetic retinopathy is needed. Deep diagnostic model has been leveraged for a variety of classification tasks including automated classification of diabetic retinopathy~\citep{gulshan2016development}. Furthermore, for the first time, the US Food and Drug Administration has approved an artificial intelligence diagnostic device that does not need a specialized doctor to interpret the results. The software program, called IDx-DR, can detect a form of eye disease by looking at photos of the retina, on April, 2018\footnote{https://www.fda.gov/news-events/press-announcements/fda-permits-marketing-artificial-intelligence-based-device-detect-certain-diabetes-related-eye}.
Gulshan \etal~\citep{gulshan2016development} presented a deep learning algorithm that is capable of interpreting signs of DR in retinal photographs, potentially helping doctors screen more patients in settings with limited resources.

The chest X-ray is among the most commonly accessible and cost-effective medical imaging examinations in medical community~\citep{world2001standardization}. It can be used for diagnosis of numerous lung ailments including atelectasis, cardiomegaly, mass, effusion and \etal~\citep{franquet2001imaging}. The ChestX-ray14 dataset released by Wang \etal~\citep{wang2017chestx} collected 112,120 frontal-view chest X-ray images that are individually labeled, with up to 14 different thoracic diseases of 30,805 unique patients. The availability of this large scale dataset makes it feasible to apply deep learning technology into this area without a need for data augmentation.
Triggered by the work of Wang \etal using convolution neural networks (CNNs) from the computer vision domain, several research groups have applied CNNs for chest X-ray classification. In~\citep{yao2017learning}, the authors presented a combination of a CNN and a recurrent neural network to exploit label dependencies. As a CNN backbone, they used a DenseNet~\citep{huang2017densely} model which was adapted and trained entirely on X-ray data. Li \etal~\citep{li2018thoracic} presented a framework for pathology classification and localization using CNNs. More recently, Rajpurkar \etal~\citep{rajpurkar2017chexnet} proposed a transfer learning strategy with fine tuning using DenseNet-121~\citep{huang2017densely}, and boost the multi-label classification performance on the ChestX-ray14 dataset. 

\subsection{Adversarial Attack}
%==>second, introduce adversarial attack on nature and medical images
Despite the successful application of deep neural networks to disease diagnosis in medical image, the discovery of so-called ``adversarial examples'' has exposed vulnerability in even state-of-the-art learning systems in machine learning community. 
%==>introduce  discovery
Szegedy \etal~\citep{szegedy2013intriguing} first discovered an intriguing weakness of deep neural networks in the context of image classification. They show that despite their high accuracies, modern deep models are surprising susceptible to adversarial attacks in the form of slightly perturbed images resembling original images, but maliciously designed to fool pre-trained models. Such attacks can cause a neural network classifier to completely change its prediction for the image. Even worse, the attacked models report high confidence on the wrong prediction.
Moreover, the same image perturbation can fool multiple network classifiers.

Since the first finding of Szegedy \etal~\citep{szegedy2013intriguing}, various approaches have been proposed for creating adversarial perturbations.
Goodfellow \etal~\citep{goodfellow2014explaining} proposed Fast Gradient Sign Method (FGSM) to generate adversarial examples. It computes the gradient of the loss function with respect to pixels, and moves a single step based on the sign of the gradient. Based on this work, Madry \etal~\citep{madry2017towards} presents an iterative algorithm to compute the adversarial perturbations by assuming that the loss function can be linearized around the current data point in each iteration, named as Projected Gradient Descend (PGD). In addition to these gradient-based attack methods, optimizing-based methods such as~\citep{poursaeed2018generative} defines a loss function based on the perturbation constrains and the pre-trained classification model's loss. Then they use the least likely class of each category as the training target with optimizer like stochastic gradient descent (SGD)~\citep{zhang2004solving} or adaptive moment estimation (Adam)~\citep{kingma2014adam} to create the perturbations. Different from gradient-based methods, the latter approach has a good generalization performance. We can use training data to learn the parameters of perturbations' distribution. During inference, we are capable of  forwarding pass from trained generative structure to generate adversarial perturbations for test samples without optimization.
Moreover, the same image perturbation can fool multiple network classifiers. The profound implications of these results triggered a wide interest of researchers in
adversarial attacks and their defenses for deep learning in general.

%===>adversarial attack on medical image 
Apart from the recent progress of adversarial attack on nature image area, medical image domain has also concerned about this topic.
\citep{paschali2018generalizability} utilized adversarial examples to evaluate the robustness of Inception of skin lesion classification and UNet of whole brain segmentation.
Taghanaki \etal~\citep{taghanaki2018vulnerability} presented several different adversarial attacks on classification of chest X-ray images and investigated how two different standard deep neural networks perform against adversarial perturbations. 
\citep{finlayson2019adversarial} hopes to highlight these vulnerabilities in medical community with the insight of healthcare domain, instead of technique domain. \citep{ma2020understanding} analyzed the different performances of medical images and natural images when attacked by adversarial perturbations, and found that medical images are more vulnerable to attack and easier to detect.

\subsection{Adversarial Defense}
Besides, several methods have also been proposed for defending against adversarial attack~\citep{cisse2017parseval,papernot2016distillation,alemi2016deep}, 
such as preprocessing techniques~\citep{guo2017countering, buckman2018thermometer}, detection algorithms~\citep{metzen2017detecting, feinman2017detecting}, and various theoretically motivated heuristics~\citep{xiao2018training, croce2018provable}, but it is maybe an ill-matched games, so far no defense strategy is safe enough.
Fawzi \etal~\citep{fawzi2018adversarial} derive fundamental upper bounds on the robustness of any classifier to perturbations, which provides a baseline to the maximal achievable robustness. When the latent space of the data distribution is in high dimension, the analysis shows that any classifier is vulnerable to very small perturbations. Their results further suggest the existence of a tight relation between robustness and linearity of the classifier in the latent space.
Shafahi \etal~\citep{shafahi2018adversarial} use well-known results from high-dimensional geometry, specifically isoperimetric inequalities, to provide bounds on the robustness of classifiers. 
These papers argue that the high dimensionality of the input space can present fundamental barriers on classifier robustness. 
	
%% %% %% ---------------------------------------------------%% %% ^_^
\section{Materials}
\label{S3}
In this section, we introduce three representative deep diagnostic medical models in detail as well as their datasets used in our study.
\subsection{Datasets}
Three public datasets are used in this study, including (1) the International Skin Imaging Collaboration (ISIC) dataset with dermoscopic image for \textbf{Melanoma} classification\footnote{https://www.isic-archive.com}; 
(2) the \textbf{Messidor} dataset with eye fundus color numerical images of the posterior pole\footnote{http://www.adcis.net/en/third-party/messidor}, 
and (3) the \textbf{ChestX-ray14} dataset with X-ray images\footnote{https://www.kaggle.com/c/ccc-chestx-ray14-multi-label-classication/data}. 

\subsection{Data Preparation}
We preprocess the images of melanoma dataset by center-cropping the images to a squared size with the length of each side equal to 0.8 $\times$ min(Height, Width), then resizing to 256 $\times$ 256 and center-cropping to 224 $\times$ 224.

Before inputting the images of ChestX-ray14 dataset into the network, we downscale the images to 224 $\times$ 224 and normalize based on the mean and standard deviation of images in the ImageNet training set. We also augment the training data with random horizontal flipping.

All Messidor data are processed following a standard pipeline, including resizing to 299 $\times$ 299, random rotating in 20, random horizontal flipping and random vertical flipping.

\subsection{Pre-trained Models}
To explore the performance of classification models more comprehensive, three types of disease diagnostic tasks are performed, including binary, multi-class, and multi-label classification. We use representative deep diagnostic models across three different medical image domains in this study. These networks are pre-trained as follows. 

(1) Pre-trained IPMI2019-AttnMel for melanoma detection~\citep{yan2019melanoma}. This network is trained end-to-end for $50$ epochs using stochastic gradient descent with momentum. The initial learning rate is $0.01$ and is decayed by $0.1$ every $10$ epochs. The code and pre-trained parameters are provided online\footnote{https://github.com/SaoYan/IPMI2019-AttnMel}.

(2) Pre-trained Inception\_v3 for detection of diabetic retinopathy~\citep{gulshan2016development}. This network is
trained for $50$ epochs using adaptive moment estimation (Adam)~\citep{kingma2014adam} with $\beta_1 = 0.9$ and $\beta_2 = 0.99$. The initial learning rate is $0.01$ and is decayed by $0.1$ every $10$ epochs. Data augmentation (\ie, random cropping, rotation, and flipping) is applied via PyTorch~\citep{paszke2017automatic} transform modules. The source code can be found online\footnote{https://github.com/mikevoets/jama16-retina-replication}.

(3) Pre-trained CheXNet for classification of $14$ types of diseases on ChestX-Ray~\citep{huang2017densely}. This network are initialized by weights from a model pre-trained on ImageNet~\citep{deng2009imagenet}, and fine-tuned using Adam with standard parameters ($\beta_1 = 0.9$ and $\beta_2 = 0.999$)~\citep{kingma2014adam}. We train the model using mini-batches of size $16$. We use an initial learning rate of $0.001$ that is decayed by a factor of $10$ each time the validation loss plateaus after an epoch, and pick the model with the lowest validation loss. The code and pre-trained parameters are available online\footnote{https://github.com/arnoweng/CheXNet}.

%%%%%%%%%%%%        New Section  :)        %%%%%%%%%%%%%%
\section{Proposed Method}
\label{S4}
\begin{figure}[t]
\setlength{\abovecaptionskip}{0pt}
\setlength{\belowcaptionskip}{0pt}
\setlength{\abovedisplayskip}{0pt}
\setlength{\belowdisplayskip}{0pt}
	\centering
	\includegraphics[width=\textwidth]{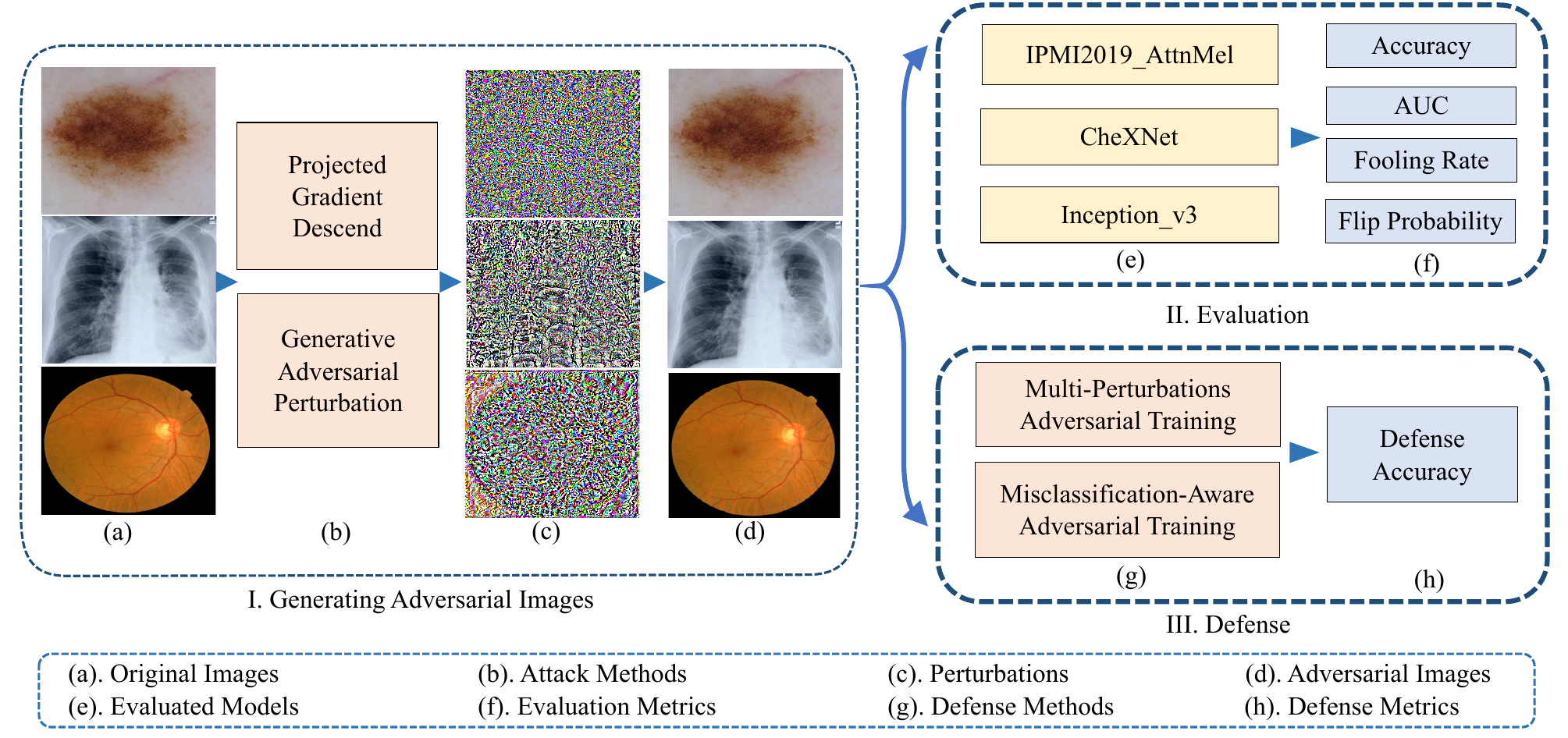}
	\caption{Overview of the proposed pipeline for evaluating the robustness of deep diagnostic models against adversarial attacks. Specifically, we first use adversarial attack methods to generate adversarial images for three datasets (Part I). Then, we evaluate the robustness of pre-trained deep diagnostic models by these adversarial images and four evaluation metrics (Part II). We further proposed two defense methods to enhance the robustness of these deep diagnostic models (Part III).}
	\label{fig:method}
\end{figure}

In this part, we introduce in detail our method for evaluating deep diagnostic models against adversarial attacks, including the method to generate adversarial image (Section \ref{sec:generative adversarial image}), a specific constraint on adversarial image (Section \ref{sec:constraint adversarial image}), the metrics (Section \ref{sec:evaluation criterion}) as well as the benchmark (Section \ref{sec:benchmark}) we used to evaluate the robustness. We further develop two new defense methods called \textit{\textbf{Multi-Perturbations Adversarial Training (MPAdvT)}} and \textit{\textbf{Misclassification-Aware Adversarial Training (MAAdvT)}} to significantly improve the robustness of deep diagnostic models (Section \ref{sec:defense}). The pipeline of our method is shown in Figure~\ref{fig:method}.

\subsection{Generating Adversarial Images}
\label{sec:generative adversarial image}
To evaluate the robustness of deep diagnostic models, we firstly need to create a new dataset as input. Here we use two adversarial attacks for this purpose, {\em i.e.} gradient-based method and optimizing-based method.

Gradient-based method generates perturbation by changing the gradient of pre-trained deep diagnostic models for each original input image. This method has a better mathematical theoretical support but no generalization performance. 

Conversely, optimizing-based method attempts to learn the distribution of perturbation with generative structure parameters. So the training data can be used to learn the parameters of perturbations' distribution. During inference, we are capable of forwarding pass from trained generative structure to generate adversarial perturbations for test samples. 

\subsection{Constraint on Adversarial Image}\label{sec:constraint adversarial image}
When generating the adversarial image, we need to add certain constraints to the perturbation so that adversarial image has invisible difference with the original one. It should be noted that the perturbation we generated is carefully designed, not a random noise. Here we have different constraint methods for gradient-based method and optimizing-based method respectively.

\subsubsection{Projected Gradient Descend}\label{sec:PGD Attack}
\begin{figure}[t]
\setlength{\abovecaptionskip}{0pt}
\setlength{\belowcaptionskip}{0pt}
\setlength{\abovedisplayskip}{0pt}
\setlength{\belowdisplayskip}{0pt}
	\centering
	\includegraphics[width=\textwidth]{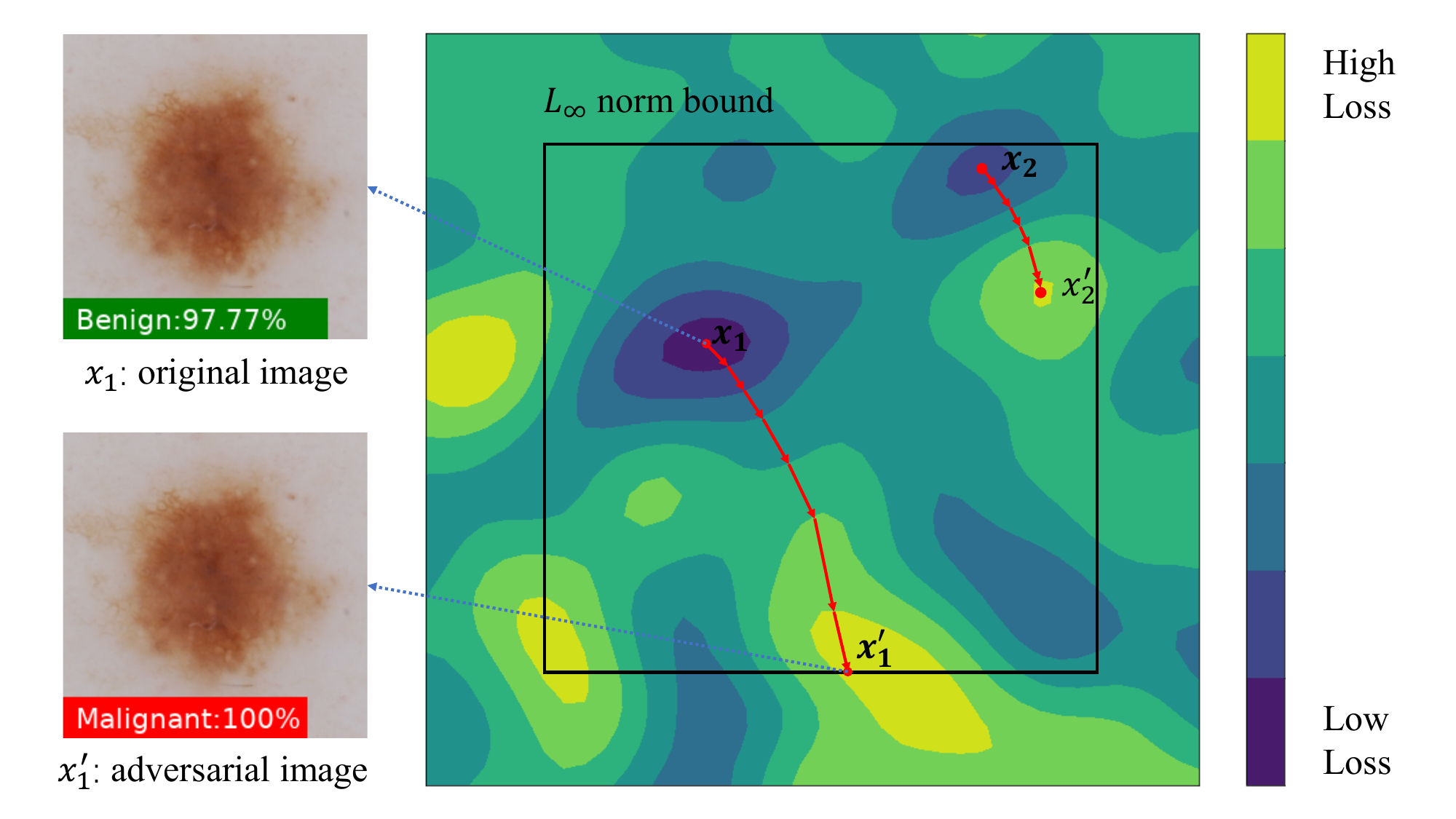}
	\caption{Changes of loss function under projected gradient descent (PGD) attack. After PGD attack, there is an adversarial example with a high loss value within the $L_\infty$ norm constraint. Here, $x_1$ and $x_2$ represent the original and clean images with low loss values, respectively, while ${x_1}'$ and ${x_2}'$ represent their corresponding adversarial images with high loss values after projected gradient descent iterations, respectively.
	}
	\label{fig:PGD}
\end{figure}

We use projected gradient descend method~\citep{madry2017towards} as gradient-based method in practice. Its constraint definition is as follows:

\noindent\textbf{Definition 1. }Let us consider a standard classification task with an underlying data distribution $\mathcal{D}$ over pairs of examples $x \in \mathbb{R}^d$ and corresponding labels $y \in \{1,2,\cdots,k\} $. We also assume that we are given a suitable loss function $\ell(x, y, \theta)$, for instance the cross-entropy loss for a neural network. As usual, $\theta \in \mathbb{R}^p$ is the set of model parameters.
The Projected Gradient Descent (PGD) on the negative loss function is defined as follows:

\begin{equation}
x_0' = x + Uniform(-\epsilon, +\epsilon) 
\end{equation}
\begin{equation}
x_{t+1} ' = Clip_{x,\epsilon}\{x_t'+\alpha \times sign(\nabla_x \ell(x, y, \theta))\}
\end{equation}
where $Uniform(\cdot)$ is a uniform function, $Clip_{x,\epsilon}\{x '\}$ is the function which performs per-pixel clipping of the image $x'$, so the result will be in $L_\infty$ norm $\epsilon$-neighbourhood of the source image $x$. $sign(\cdot)$ is an odd mathematical function that extracts the sign of a real number. $\alpha$ is step size for each attack iteration. $t$ is iteration number. 

As shown in Figure~\ref{fig:PGD}, we perform the following steps in the PGD attack: 1) Starting from a random perturbation in the $L_\infty$ norm constraint around a sample $x_1$; 2) taking a gradient iteration step in the positive direction of greatest loss; 3) projecting perturbation back into $L_\infty$ norm constraint if necessary; 4) repeating 2) - 3) until convergence. 
 The iteration of sample $x_1$ describes that perturbation has reached the norm constraint but not convergence. In this situation, the iteration stops and the adversarial image ${x_1}'$ with imperceptible changes is produced. The sample $x_2$ shows the perturbation has converged to the local maximum of the loss before reaching the norm constraint. 

\subsubsection{Generative Adversarial Perturbations}

We use Generative Adversarial Perturbations (GAP) method~\citep{poursaeed2018generative} as our optimizing-based method in practice, the training architecture of GAP attack is shown in Figure~\ref{fig:GAP}. Its constraint definition is as follows:

\begin{figure}[t]
	\setlength{\abovecaptionskip}{0pt}
	\setlength{\belowcaptionskip}{0pt}
	\setlength{\abovedisplayskip}{0pt}
	\setlength{\belowdisplayskip}{0pt}
	\centering
	\includegraphics[width=\textwidth]{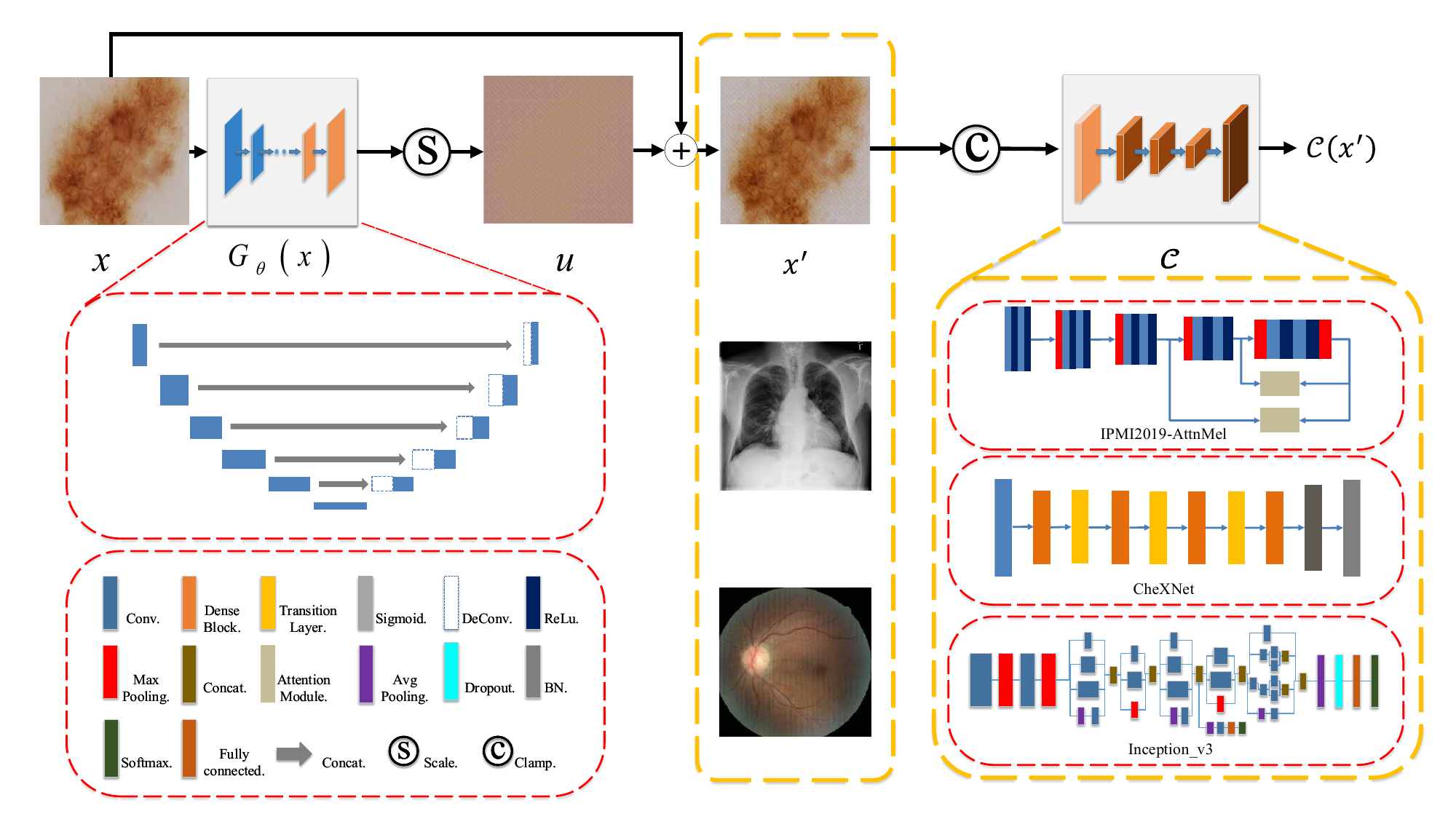}
	\caption{Training architecture for generating adversarial perturbations. The generator $G_{\theta}(x)$ outputs a perturbation $u$, which is scaled to satisfy a norm constraint. It is then added to the original image, and clipped to produce the perturbed image $x'$. Finally, we update the generator’s parameters with the loss function calculated by the outputs of pre-trained classification model $\mathcal{C}$. We use the U-Net architecture as generator, and three models (\ie, IPMI2019-AttnMel, CheXNet and Inception\_v3) as the pre-trained classification module in this work.
	}
	\label{fig:GAP}
\end{figure}

\noindent\textbf{Definition 2. }Let $\mathcal{D}$ denote the distribution of input images $x$ in $\mathbb{R}^d$ and $y \in \{1,2,\cdots,k\} $ denotes the corresponding labels, $\mathcal{C}$ be a pre-trained classification model achieving high accuracy on distribution $\mathcal{D}$. The GAP attack aims to construct a generator $G(\cdot)$ which can produce perturbation that transforms the original image $x$ to an adversarial image $x'$, so the generator $G(\cdot)$ should satisfy:
\begin{equation}
\forall x\in \mathcal{D}, \quad
\mathcal{C}(G_{\theta}(x)+ x) \ne \mathcal{C}(x) \quad {\rm{s.t.}} \quad G_{\theta}(x) \le \epsilon
\label{id_constrain}
\end{equation}
where $\theta$ is the parameter of generator $G(\cdot)$. We require that adversarial image $G_{\theta}(x) +x $ looks similar to original image $x$. Hence, the upper bound $\epsilon$ of $G_{\theta}(x)$ should be small enough. Note that for each image $x \in \mathcal{D}$, there is a corresponding perturbation in this case.
After generator $G_\theta(x)$ outputting a perturbation, we scale it to satisfy a norm constraint, more specifically, we multiply it by min$(1,\frac{\epsilon}{\parallel G_\theta(x) \parallel_p})$ where $\epsilon$ is the upper bound of $L_p$ norm, we use $L_\infty$ norm in our study.

Then we add the original image $x$ to the generated perturbation and clip it for producing adversarial sample $x'$. We feed $x'$ to the pre-trained network $\mathcal{C}$ where we use IPMI2019-AttnMel, CheXNet and Inception\_v3 to obtain the output probability $\mathcal{C}(x')$. 
Let $\mathbb{O}(y)$ denote as one-hot encoding of label $y$ and $\mathcal{C}(x')$ as the output probability of adversarial sample $x'$. For non-target attack that do not specify a network output label, the prediction of $x'$ is expected to be different from label $y$, so we define the loss function as follows:
\begin{equation}\label{equ:GAP loss function}
l(\theta)=\frac{1}{m}\sum\limits_{i=1}^m \sum\limits_{j=1}^k\mathbb{O}_j(y_i)\times\log({\mathcal{C}_j(x'_i)})
\end{equation}
where $\theta$ denotes the parameters of generator, $m$ is the number of samples, $k$ is the number of categories, $y_i$ is the corresponding label of $x_i$, $\mathcal{C}_j(x_i')$ represents the probability of sample $x_i'$ belonging to class $j$. That is, when $l(\theta)$ becomes smaller during training, the output probability of sample $x'$ belonging to its true label will also be smaller in order to attack the network.

\subsection{Evaluation Metrics}\label{sec:evaluation criterion}
After generating the adversarial images and constraining them, we input them into the pre-trained classification models. We use binary, multi-class, and multi-label models in our study to get a more comprehensive assessment of the deep diagnostic models' robustness.

The evaluation metrics could be divided into four components. 
The first component we evaluate the models' robustness by comparing the accuracy and fooling ratio between adversarial image and original image.
The second component we visualize the intermediate feature changing of the model.
In the third component, we analyze the discriminability of learned features with the increase of network layers. 
In the last component, we analyze the correlation of labels between adversarial image and original one.

\subsection{Benchmark for Common Perturbation Robustness Evaluation}\label{sec:benchmark}
\begin{figure}[t]
\setlength{\abovecaptionskip}{0pt}
\setlength{\belowcaptionskip}{0pt}
\setlength{\abovedisplayskip}{0pt}
\setlength{\belowdisplayskip}{0pt}

	\centering
	\includegraphics[width=\textwidth]{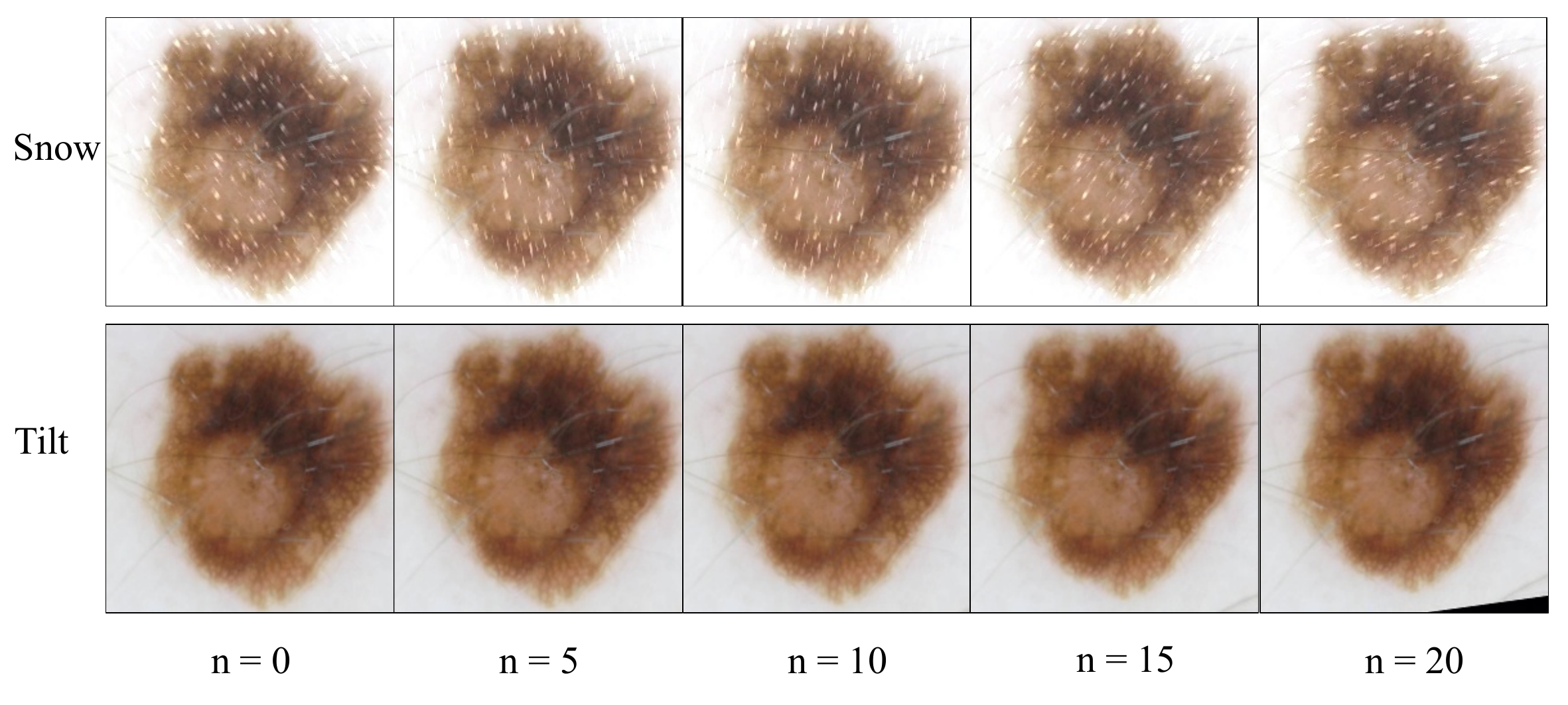}
	\caption{Example frames from the beginning (n = 0) to end (n = 20) of some Tilt and Snow perturbation sequences in the proposed Robust-Benchmark. 
	}
	\label{fig:benchmark}
\end{figure}

In addition to evaluating the robustness of deep diagnostic models against adversarial attack, we further evaluate the robustness of the model against common perturbations.
Models lacking in perturbation robustness produce erratic predictions which undermines user trust.
\citep{dan2019benchmarking} proposes new datasets called IMAGENET-C and  IMAGENET-P which enable researchers to benchmark a classifier’s robustness to common natural image corruptions and perturbations. These datasets have a connection with adversarial distortions and play a key role in safety-critical applications and are widely used nowadays.
Following~\citep{dan2019benchmarking}, we create a new dataset (called \emph{Robust-Benchmark}) of medical images to evaluate the common perturbation robustness of deep diagnostic models in a standard way.
We hope \emph{Robust-Benchmark} will serve as a general dataset for benchmarking robustness to image perturbations.

\textbf{Design of \emph{Robust-Benchmark}.}
The \emph{Robust-Benchmark} consists of 14 diverse perturbations types (\ie, {\em Brightness, Gaussian blur, Gaussian noise, Motion blur, Rotate, Scale, Shear, Shot noise, Snow, Spatter, Speckle noise, Tilt, Translate, Zoom blur}) applied to test images of three datasets in our study. The perturbations are drawn from four main categories (\ie, {\em noise, blur, weather, and digital}). Each perturbation type has five levels of severity since perturbations can manifest themselves at varying intensities.
This dataset is a variant of the original test set, containing a number of image sequences. Specifically, each image sequence begins with the original clean image, and the following frames are created by adding a type of perturbation ({\em e.g.}, noise, blur, weather or digital distortions) to the original image. Examples are shown in Figure~\ref{fig:benchmark}, we can observe that the following frame is merely different with the former one for it just adding small perturbation.
We evaluate the perturbation robustness of three deep diagnostic models with their \emph{Robust-Benchmark} datasets, respectively. Note that networks should be trained on their original datasets rather than \emph{Robust-Benchmark} because \emph{Robust-Benchmark} dataset is just an evaluation benchmark.

\textbf{Evaluation Metric.}
Then, we evaluate the robustness of three models to common perturbations on the {\emph {Robust-Benchmark}} dataset. 
The \textbf{Flip Probability (FP)} is used as an evaluation metric because it has been proven to effectively measure the robustness of a model~\citep{dan2019benchmarking,xie2020self,kamann2020benchmarking,xie2020adversarial}.
Denote $r$ perturbation sequences as $\mathcal{S}=\{(x_1^{(i)},x_2^{(i)},\cdots,x_n^{(i)})\}_{i=1}^{r}$. The FP value of a network $\mathcal{C}$ on perturbation sequences $\mathcal{S}$ is defined as:
\begin{equation}
\small
FP^\mathcal{C}=\frac{1}{r(n-1)}\sum_{i=1}^{r}\sum_{j=2}^{n}I \left(\mathcal{C}(x_j^{(i)}) \neq \mathcal{C}(x_{j-1}^{(i)}) \right),
\label{eq: fp}
\end{equation}
where $r$ is the number of perturbations, $n$ is the number of frames of each noise sequence. For noise perturbation sequences, which are not temporally related, $x_1^{(i)}$ is clean and $x_j^{(i)} (j\textgreater 1)$ is perturbed image of $x_1^{(i)}$. And $I(\cdot)$ is an indicator function, where $I(\cdot)=1$ if $\mathcal{C}(x_j^{(i)})\neq \mathcal{C}(x_{j-1}^{(i)})$, and $I(\cdot)=0$ otherwise. %satisfies, function $I$ returns 1, otherwise returns 0.
With Equation~(\ref{eq: fp}), a larger FP value denotes that the corresponding model is more vulnerable to noise perturbations, and vice versa.

\subsection{Defense Method for Improving Robustness}
\label{sec:defense}
Several defense techniques have been proposed to make deep neural networks (DNNs) more robust to adversarial examples, including defensive distillation~\citep{papernot2016distillation}, gradient regularization~\citep{gu2014towards,papernot2017practical,ross2018improving}, model compression~\citep{liu2018security}, adversarial denoising~\citep{xie2019feature}, among which adversarial training has been demonstrated to be the most effective~\citep{athalye2018obfuscated}.
Adversarial training can be regarded as a data augmentation technique that trains DNNs on adversarial examples, and can be viewed as solving the following min-max optimization problem~\citep{madry2017towards}:
\begin{equation}
\min_\theta \frac{1}{m} \sum_{i=1}^{m} \max_{||x_i' -x_i||_p \le \epsilon} \ell(\mathcal{C}_\theta(x_i'),y_i)
\end{equation}
where $m$ denotes the number of training examples, $x_i'$ is an adversarial example of the original image $x_i$, $\mathcal{C}(\cdot)$ is the classification model and $\ell(\cdot)$ is the classification loss function. The inner maximization is used to generate adversarial images, which are employed as training set to train the robust classification model in outer minimization.
Recently, adversarial training with adversarial examples generated by Projected Gradient Descent (PGD)~\citep{madry2017towards} has been demonstrated to be the most effective method that can train moderately robust DNNs without being fully attacked~\citep{athalye2018obfuscated}. 

\begin{algorithm} [t]
	\small
	\label{alg:MPAdvT}
	\small
		\caption{Multi-Perturbations Adversarial Training (MPAdvT)}  
		\LinesNumbered
		\KwIn{
			Training data $\{x_i,y_i\}_{i=1}^{m}$, outer iteration epoch $T_O$, inner iteration step $T_I$, maximum perturbation $\epsilon$, step size for inner optimization $\alpha_I$, step size for outer optimization $\alpha_O$
		}
		
		\textbf{Initialize:} Standard random initialization of $\mathcal{C}_\theta$
		
		\For{$s = 1,\cdots,T_O$}{
			Uniformly sample a minibatch of training data $B^{(s)}$ 
			
			$p \leftarrow \mathcal{U}(0,1)$, where $\mathcal{U}$ is a uniform distribution 
			
			$\epsilon \leftarrow \mathcal{U}(0.01,0.04) $ 
			
			$T_I \leftarrow \mathcal{U}(1,5)$
			
			\For{$x_i\in B^{(s)}$}{
				\If{$p \ge 0.5$}{
					$x_i = x_i + \mathcal{U}(-\epsilon,+\epsilon)$
					
					\For{$t = 1,\cdots,T_I$}{
						$x_i \leftarrow Clip_{x_i,\epsilon}(x_i+\alpha_I \times sign(\nabla_{x_i} \ell (\theta,x_i,y_i)))$
					}
				}
				$x_i ' \leftarrow x_i$
				
				$\theta \leftarrow \theta- \alpha_O \sum_{x_i\in B^{(s)}} \nabla_\theta \ell(\theta,x_i ',y_i)$
			}
		}
		\KwOut{Robustness classifier $\mathcal{C}_\theta$}
		
\end{algorithm}

However, adversarial training attempts to minimize the maximum loss within a fixed-size neighborhood of the training data generated by PGD attack~\citep{ding2018max}. Despite advancements made in recent years~\citep{hendrycks2019using,zhang2019you,shafahi2019adversarial,stanforth2019labels}, a fundamental problem in adversarial training is that the perturbation level $\epsilon$ has to be set in advance and is fixed throughout the training process.
If $\epsilon$ is set too small, the resulting model lacks robustness, if too large, the resulting model lacks accuracy. 
To remedy this problem, we propose
\textit{\textbf{Multi-Perturbations Adversarial Training (MPAdvT)}} that trains deep diagnostic models with different perturbation levels $\epsilon$ and different iteration steps $t$ during the training process. The detailed training procedure of MPAdvT is described in Algorithm~\ref{alg:MPAdvT}.

Recall that the formal definition of an adversarial example is conditioned on it being correctly classified~\citep{carlini2019evaluating}. From this perspective, there is no definition on adversarial examples generated from misclassified examples. Most recent adversarial training variants neglect this problem and treat all examples equally in the adversarial training process.
The influence of misclassified and correctly classified examples on the final robustness of adversarial training has not been payed sufficient attention. Wang~\etal~\citep{wang2020improving} find that the manipulation on misclassified examples has more impact on the final robustness, and the minimization techniques are more crucial than maximization ones under the min-max optimization framework in natural image field. Motivated by~\citep{wang2020improving}, for a $k$-class  ($k\ge 2$) classification task, we add a \textit{\textbf{misclassification aware regularization}} to adversarial loss function. For these misclassified examples, it is hard to minimize the standard adversarial loss directly, as themselves cannot be classified correctly, even without any perturbations. So we use Kullback–Leibler (KL) divergence to encourage the output of classifier to be stable against misclassified adversarial examples. 
Let $\mathcal{D}$ denote the distribution of input images $x$ in $\mathbb{R}^d$ and $y \in \{1,2,\cdots,k\} $ denote the corresponding labels, we have
\begin{equation}
KL(\mathcal{C}(x_i)||\mathcal{C}{(x_i')}) =  \sum_{j=1}^{k}\mathcal{C}_j(x_i)log\frac{\mathcal{C}_j(x_i)}{\mathcal{C}_j(x_i')}
\end{equation}

where $x_i'$ is the adversarial image of original image $x_i$, $\mathcal{C}_j(x_i)$ represents the probability of $x_i$ belonging to class $j$ outputted by classification model $\mathcal{C}$. It reflects the different output distribution between adversarial image and original image. Then the regularization term is defined as follows:
\begin{equation}
\mathcal{R}_i(\theta)= KL(\mathcal{C}(x_i)||\mathcal{C}(x_i')) \times(1-\mathcal{C}_{y_i}(x_i))
\end{equation} 
$1-\mathcal{C}_{y_i}(x_i)$ emphasizes learning on misclassified examples, this will be large for misclassified examples and small for correctly classified examples.

Based on this misclassification aware regularization, we further propose the \textit{\textbf{ Misclassification-Aware Adversarial Training (MAAdvT)}} with the loss function
\begin{equation}
\mathcal{L}^{MAAdvT}(\theta)=\frac{1}{m}\sum_{i=1}^m\ell(x_i,y_i,\theta)
\end{equation}
where $\ell(x_i,y_i,\theta)$ is defined as
\begin{equation}
\ell(x_i,y_i,\theta):=CE(\mathcal{C}(x_i'),y_i) + \lambda \mathcal{R}_i(\theta)
\end{equation}
$CE(\cdot)$ is the Cross-Entropy loss, $\lambda$ is a tunable scaling parameter that balances the two parts of the final loss, and is fixed for all training examples. The detailed training procedure of MAAdvT is described in Algorithm~\ref{alg:MAAdvT}.

\begin{algorithm}
\small
\label{alg:MAAdvT}
\caption{Misclassification-Aware Adversarial Training (MAAdvT)}
\KwIn{Training data $\{x_i,y_i\}_{i=1}^{m}$, outer iteration epoch $T_O$, inner iteration step $T_I$, maximum perturbation $\epsilon$, step size for inner optimization $\alpha_I$, step size for outer optimization $\alpha_O$, tunable scaling parameter $\lambda$}

\textbf{Initialize:} Standard random initialization of $\mathcal{C}_\theta$

\For{$s = 1,\cdots, T_O$}{
	Uniformly sample a minibatch of training data $B^{(s)}$ \newline
	\For{$x_i\in B^{(s)}$}{
		$x_i'\leftarrow PGD(x_i, y_i, \epsilon, \alpha_I,T_I)$ \quad \# $PGD(\cdot)$ is PGD attack
	
		$\mathcal{R}_i(\theta)\leftarrow KL(\mathcal{C}(x_i)||\mathcal{C}(x_i')) \times(1-\mathcal{C}_{y_i}(x_i))$

        $\mathcal{L}_i^{MAAdvT}(\theta)\leftarrow CE(\mathcal{C}(x_i'),y_i) + \lambda \mathcal{R}_i(\theta)$
        
        $\theta \leftarrow \theta - \alpha_O \sum_{x_i\in B^{(s)}} \nabla_\theta \mathcal{L}_i^{MAAdvT}(\theta)$
 }
}
\KwOut{Robust classifier $\mathcal{C}_\theta$}
\end{algorithm}

%%%%%%%%%%%%        New Section  :)        %%%%%%%%%%%%%%

\section{Experiment}
\label{S5}
In this section, we first introduce experimental settings (Section~\ref{sec:experiment setting}) including the parameters we used in the experiments.
Then we present the experimental results of three models under adversarial attacks from two aspects. The first part (Section~\ref{sec:single-label}) is based on single-label classification problems (with each image only annotated by one single label) and two deep diagnostic models ({\em i.e.}, IPMI2019-AttnMel for binary classification and Inception\_v3 for multi-class classification). We analyze the change of classification results and intermediate results of feature extraction of two models under adversarial attacks. The second part (Section~\ref{multi-label}) is based on multi-label classification problems (with each image annotated by multiple class labels) and the CheXNet model. 
We also show the Flip Probability (FP) of three models when evaluated by \emph{Robust-Benchmark} datasets respectively (Section~\ref{sec:benchmark Results}).
Finally, we compare the robustness between natural model and defense model trained with our MPAdvT and MAAdvT, the results indicate that the robustness of deep diagnostic models against adversarial attacks can be significantly improved by the use of defense methods (Section~\ref{sec:defense results}).
The code and trained models can be found online\footnote{https://github.com/MengtingXu1203/EvaluatingRobustness}.

\subsection{Experimental Setting}\label{sec:experiment setting}
As for PGD attack, in Component 1, the perturbation level $\epsilon$ are set to $2.0/255$ and $4.0/255$ with the iteration steps to $1.0$ and $4.0$, respectively. In Component 2, the $\epsilon$ is $5.0/255$ and the iteration step is $40.0$. In Component 3, $\epsilon$ constraint and iteration step are $10/255$ and $60$. In Component 4, the $\epsilon$ constraint is $4.0/255$ and the iteration step is $4.0$. With larger $\epsilon$ and more iterations, we will obtain more obvious attack effect.

For GAP attack, we use U-Net architecture~\citep {ronneberger2015u} as our perturbation generator $G$.
For parameter setting, in Component 1, the $L_\infty$ norm are set to $7$ and $11$. In Component 2, the $L_\infty$ norm constraint is set to $13$.

\subsection{Single-label Classification}\label{sec:single-label}

%%%%%%%%%%%%%-------  New Section -------%%%%%%
\noindent\textbf{Component 1: Quantitative Results}\\ 
First, we present our evaluation results by showing numerical changes in classification results. The adversarial examples are generated by projected gradient descent (PGD)~\citep{madry2017towards} and generative adversarial perturbations (GAP)~\citep{poursaeed2018generative} attacks for IPMI2019-AttnMel and Inception\_v3, respectively. 

(1) \textbf{Results under PGD Attack}. 
As shown in Table~\ref{table:adversarial_result}, the sharp decrease of accuracy (ACC) and area under receiver operating characteristic (AUC) of two models indicate that these models are vulnerable to adversarial perturbations. 
For example, with the perturbation level $\epsilon=4.0$ and iteration step $t=4.0$ for IPMI2019-AttnMel, the ACC value of identifying Melanoma drops from $87.5\%$ to $0.0\%$.
Also, under the PGD attack, the AUC of Inception\_v3 in identifying Diabetic Retinopathy decreases from $0.971$ to $0.263$. 
These results indicate that the performance of these two deep diagnostic models is poor when facing the adversarial perturbations.
Besides, we show two adversarial examples of Melanoma and Messidor in Figure~\ref{fig:adversarial_result}(a). This figure suggests, even though only a small perturbation is added to the original image, the probability scores output by IPMI2019-AttnMel and Inception\_v3 change greatly. 
\begin{table*}[t]
	\setlength{\abovecaptionskip}{5pt}
	\setlength{\belowcaptionskip}{10pt}
	\centering
	\footnotesize
	\caption{The change of numerical results achieved by IPMI2019-AttnMel for Melanoma classification and Inception\_v3 for Messidor identification. They are both attacked by PGD attack and GAP attack. ACC: accuracy; AUC: area under receiver operating characteristic; FR: fooling rate. X: training set; val: validation set; $\epsilon$: perturbation level; {\color{blue}$t$}: iteration steps.} 
	
	\centering  
	\resizebox{1\textwidth}{0.11\textheight}
	{
		\begin{tabular}{cc c cc  c cccc}
			
			\toprule
			\multicolumn{2}{c}{\multirow{3}{*}{Model}} & \multirow{3}{*}{Original/Clean Images}  & \multicolumn{2}{c} {{Images with PGD Attack}}&  &\multicolumn{4}{c} {{Images with GAP Attack}}  \\  
			\cmidrule{4-5} \cmidrule{7-10}
			
			& && $\epsilon = 2.0/255$ & $\epsilon=4.0/255$ &  & \multicolumn{2}{c}{$L_\infty = 7$} & \multicolumn{2}{c}{$L_\infty = 11$}
			\\
			& && $t = 1.0$ & $t = 4.0$ &  & X &val. & X& val.
			\\
			
			\midrule
			\specialrule{0em}{1pt}{1pt}
			\multirow{3}{*}{{IPMI2019-AttnMel}}%\begin{sideways}\textbf{Melanoma}\end{sideways}}
			%			\specialrule{0em}{10pt}{10pt}
			&ACC & $87.5$\% &10.8\% & 0.0\%&&31.1\%&30.1\%&21.7\%& 21.9\% \\
			\specialrule{0em}{3pt}{3pt}
			&AUC & 0.744 &0.067& 0.000&&0.558&0.539&0.514&0.493 \\
			\specialrule{0em}{3pt}{3pt}
			&FR & - &74.7\%& 85.5\%&&68.9\%&71.2\% & 78.3\%& 80.5\% \\
			\specialrule{0em}{1pt}{1pt}
			\hline	
			\specialrule{0em}{1pt}{1pt}
			\multirow{3}{*}{{Inception\_v3}}%\begin{sideways}\textbf{Messidor}\end{sideways}}
			%			\specialrule{0em}{10pt}{10pt}
			&ACC & 89.0\% &19.0\% & 0.0\%&&37.7\%&29.5\%&20.7\%& 23.0\% \\
			\specialrule{0em}{3pt}{3pt}
			&AUC & 0.971 &0.715& 0.263&&0.663&0.540&0.594&0.487 \\
			\specialrule{0em}{3pt}{3pt}
			&FR & - &71.5\%& 90.5\%&&62.3\%&69.0\% & 79.6\%& 77.0\% \\
			\specialrule{0em}{0pt}{0pt}
			
			\bottomrule
		\end{tabular}
	}
	\label{table:adversarial_result}
\end{table*}

\begin{figure}[t]
	\setlength{\abovecaptionskip}{0pt}
	\setlength{\belowcaptionskip}{0pt}
	\setlength{\abovedisplayskip}{0pt}
	\setlength{\belowdisplayskip}{0pt}
	\centering
	\includegraphics[width=\textwidth]{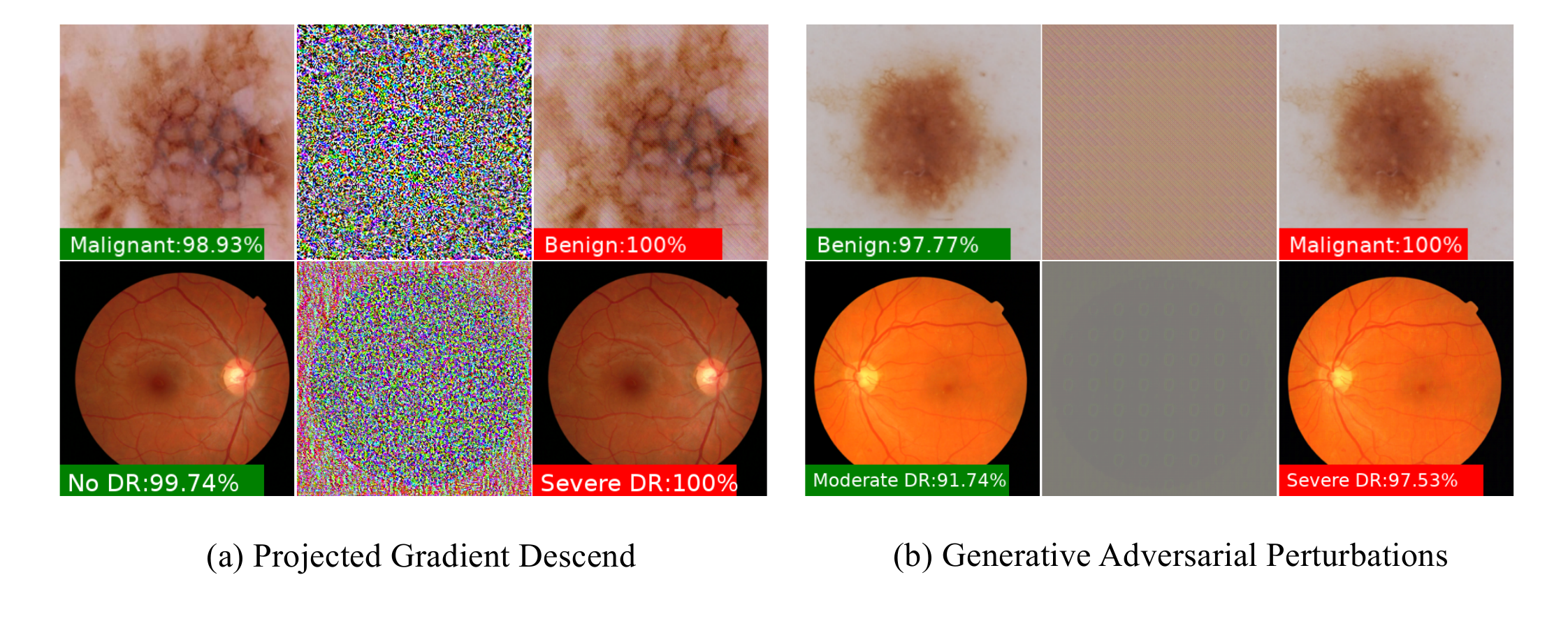}
	\caption{Visualization of original images and their corresponding adversarial examples with Melanoma (1st row) and Messidor (2nd row). The images in (a) are attacked by the projected gradient descent (PGD) perturbations, and the images in (b) are attacked by the generative adversarial perturbations (GAP). In each sub-figure, the first column shows the original image and its probability score output by a specific diagnostic model, the second column shows the adversarial perturbations, and the third column denotes the corresponding adversarial image and the output probability score of the diagnostic model. Green denotes the correct label with its probability score for the original image and red is the probability for the attacked image with its probability score. The perturbations are rescaled to $[0,255]$ for visualization.}
	\label{fig:adversarial_result}
\end{figure}

(2) \textbf{Results under GAP Attack}. For two models under the GAP attack, we can get the same conclusion that the ACC and AUC results change significantly, as shown in Table~\ref{table:adversarial_result} and Figure~\ref{fig:adversarial_result}(b). Besides, in terms of fooling rate (FR), it can be seen that perturbation has a greater impact on the single-class classifier ({\em i.e.}, IPMI2019-AttnMel) than the multi-class model ({\em i.e.}, Inception\_v3). For instance, with the upper bound $L_\infty=7$ , the FR of IPMI2019-AttnMel in binary classification of Melanoma images on the validation set is $71.2\%$, which is much higher than that ($69.0\%$) of Inception\_v3 in multi-class classification of Messidor images. It is also obvious that two models become more unreliable (with higher FR values) with the increasing of $L_\infty$ norm. 
Two adversarial examples of Melanoma and Messidor with the GAP attack are shown in Figure~\ref{fig:adversarial_result}(b), from which we can see that even if the perturbations are not visible to the human eye, the probability scores generated by two networks change greatly. This suggests that two deep diagnostic models are not robust to both PGD and GAP attacks.

\begin{figure*}[!t]
\setlength{\abovecaptionskip}{0pt}
\setlength{\belowcaptionskip}{0pt}
\setlength{\abovedisplayskip}{0pt}
\setlength{\belowdisplayskip}{0pt}
	\centering
	\includegraphics[width=1.04\textwidth]{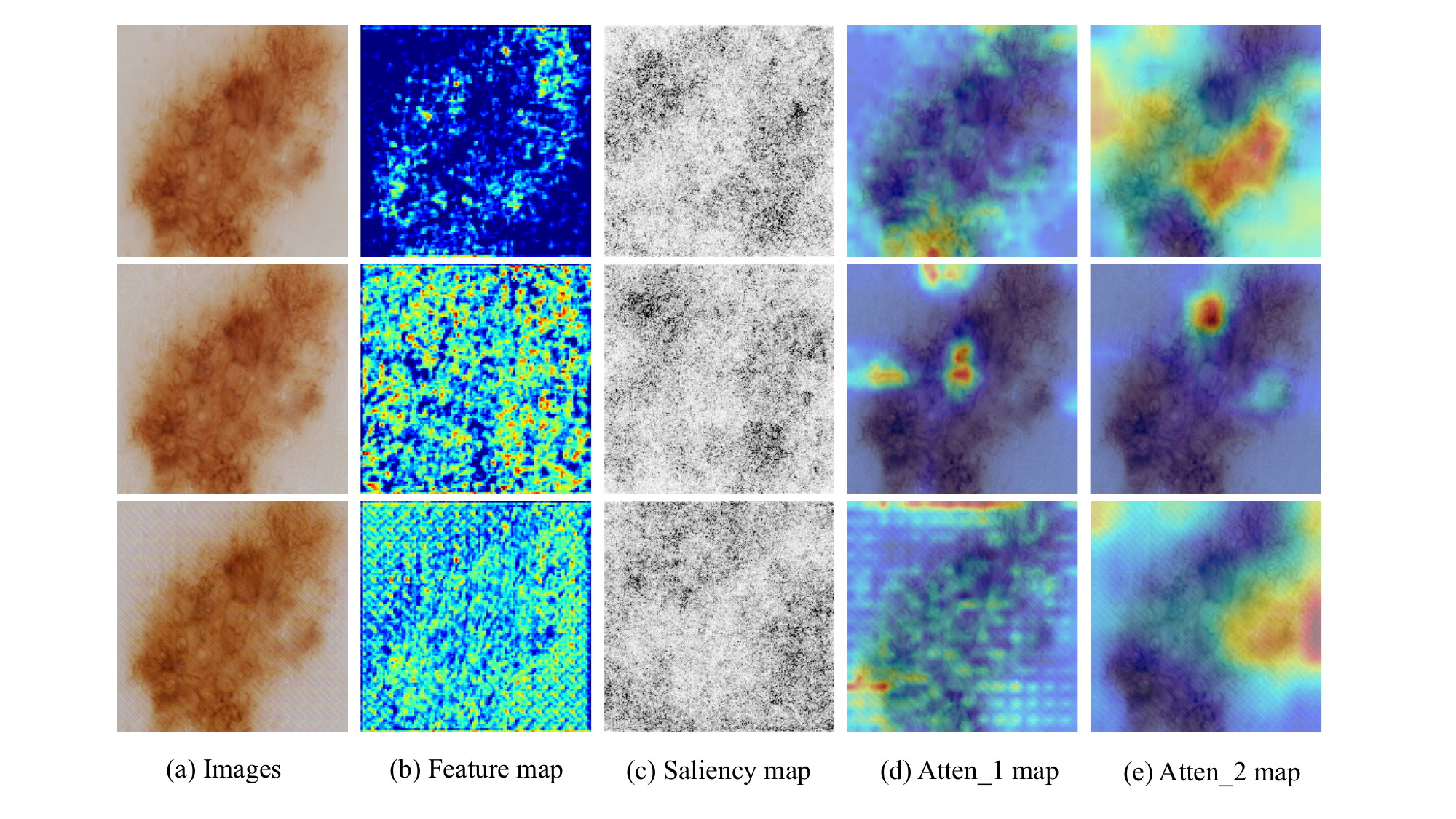}
	\caption{Visualization of intermediate features learned by the IPMI2019-AttnMel model for an original/clean image and two adversarial images. (a) shows the original/clean Melanoma image (top) and adversarial images attacked by PGD (middle) and GAP (bottom). (b) shows the feature maps at the `feature(13)\_relu' layer of the network. (c) shows the saliency maps. The attention maps derived from the `attention module1' and `attention module2' layers of the network are illustrated in (d) and (e), respectively.}
	\label{fig:visulization_result}
\end{figure*}
%%%%%%%%%%%%%-------  New Section -------%%%%%%%%%%%%%
\noindent\textbf{Component 2: Change of Intermediate Features} 

\noindent In Component 1, we show that deep diagnostic models dramatically change their outputs when attacked by adversarial perturbations. To understand why they produce different outputs under perturbations, we further study the intermediate features derived from inner layers of each network by visualizing their feature maps, saliency maps, and attention maps.

(1) \textbf{Feature Maps}.
The feature map is a mapping of where a certain kind of feature is found in the image. A high activation in a feature map means a certain feature is found/extracted from the input image. In Figure~\ref{fig:visulization_result}(a), we show an original/clean Melanoma image (top) and its adversarial images attacked by PGD (middle) and GAP (bottom). We further visualize their feature maps derived from the `feature(13)\_relu' layer in IPMI2019-AttnMel in Figure~\ref{fig:visulization_result}(b).
From Figure~\ref{fig:visulization_result}(a)-(b), one can observe that the learned features for the clean image focus on semantically informative regions (represented in red), while the features of the two adversarial images are activated globally (without any specific focus).

(2) \textbf{Saliency Maps}.
The saliency map of an input image highlights regions that cause the model output to change the most, based on the gradients of the classification loss with respect to the input.
For each pixel in the input image, this gradient tells us how correctly the score changes when the pixel changes slightly. That is, the saliency map is a visualization technique to capture the pixels which the classification model really be guided by~\citep{simonyan2013deep}.
In Figure~\ref{fig:visulization_result}(c), we show the saliency maps of IPMI2019-AttnMel for the clean image (top) and two adversarial images with PGD (middle) and GAP (bottom). Figure~\ref{fig:visulization_result}(c) suggests that the saliency maps of the two adversarial images are different from that of the original/clean image.
The possible reason is that adversarial perturbations can easily change the gradients of the loss function for network training, thus guiding the network to focus on those regions that are not useful for the classification task.

(3) \textbf{Attention Maps}.
The attention modules of IPMI2019-AttnMel, which are learned together with other network parameters, estimate attention maps that highlight image regions of interest that are relevant to lesion classification. These attention maps provide a more interpretable output as opposed to only outputting a class label. For example, when diagnosing melanoma, dermatologists mainly focus more on the lesion rather than irrelevant areas such as background or hair. To imitate this visual exploration pattern, two attention modules are used in IPMI2019-AttnMel to estimate a spatial (pixel-wise) attention map~\citep{yan2019melanoma}.
Then we can efficiently utilize prior information via regularizing the attention maps with regions of interest (ROIs). With prior information, the learned attention maps are refined and the classification performance is improved. So the attention map is a training mechanism to make the classification model have higher accuracy.
In Figure \ref{fig:visulization_result}(d)-(e), we visualize attention maps of two different attention modules in IPMI2019-AttnMel for a clean image (top) and two adversarial image with PGD (middle) and GAP (bottom). From these figures, we can observe the most discriminative regions derived by IPMI2019-AttnMel on the clean image are heavily disrupted by adversarial perturbations. Specifically, the attentions are shifted from the lesion regions (see those denoted as red in the top of Figure~\ref{fig:visulization_result}(e)) to regions that are completely irrelevant to the lesion diagnosis (see those denoted as red in the middle and bottom of Figure~\ref{fig:visulization_result}(e)). 
This could imply that subtle perturbations to medical images can result in fundamentally different deep features and easily change the output probabilities of a deep diagnostic model.

\textbf{Discussion:} 
According to the changes of saliency maps that display the gradients, it may be necessary to add a regularization to smooth the loss function for robust defenses against adversarial attack. What’s more, even though networks with attention modules could bring better prediction performance, they are more vulnerable to adversarial attacks. This reminds us that when constructing deep diagnostic models, we should seriously consider the tradeoff between accuracy and robustness.

\begin{figure}[!t]
\setlength{\abovecaptionskip}{0pt}
\setlength{\belowcaptionskip}{0pt}
\setlength{\abovedisplayskip}{0pt}
\setlength{\belowdisplayskip}{0pt}
	\centering
	\includegraphics[width=\textwidth]{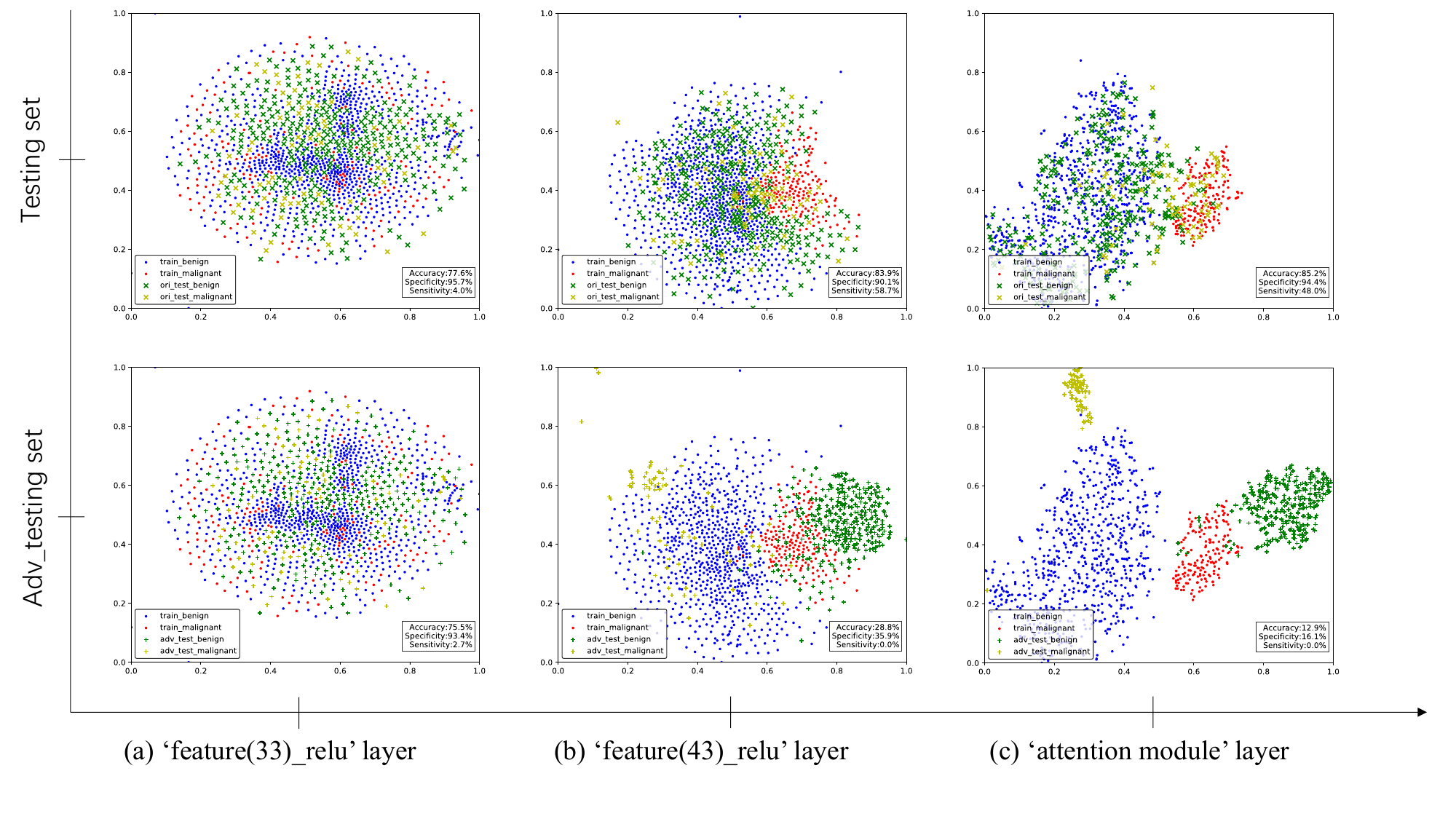}
	\caption{Visualization of 2D embeddings of features derived from IPMI2019-AttenMel for training images, original/clean test images and their corresponding adversarial test images with the PGD attack via t-SNE~\citep{maaten2008visualizing}. Feature are extracted form (a) `feature(33)\_relu' layer, (b) `feature(43)\_relu' layer, and (c) `attention module' layer of IPMI2019-AttenMel. The first row shows the original test data and the second row denotes the adversarial test data. The {\color{blue}`$\bullet$'} and `{\color{red}{$\bullet$}}' represent training data with benign and malignant respectively, which act as a control group of feature distribution. The `{\color{green}\textbf{x}}' and `{\color{yellow} \textbf{x}}' represent the distributions of original test data with label of benign and malignant respectively in the first row. The `{\color{green}\textbf{+}}' and `{\color{yellow}\textbf{+}}' represent the distributions of adversarial test data with label of benign and malignant respectively in the second row.}
	\label{fig:t-SNE}
\end{figure}
%%%%%%%%%%%%%-------  New Section -------%%%%%%%%%%%%%
\noindent\textbf{Component 3: Discriminability of Learned Features}

\noindent Through the above two components, we studied the performance of the deep diagnostic models under adversarial attacks.
We have found that these networks are prone to output erroneous results, and also the feature representations produced by their internal layers have changed (even if there is a mechanism of attention).
We now study how adversarial perturbations work at different layers of the network, by investigating the discriminative power of their learned features.% learned by the network. 

Using t-SNE~\citep{maaten2008visualizing}, we visualize the 2D embeddings of the features learned by different layers in IPMI2019-AttenMel. 
In Figure~\ref{fig:t-SNE}, the original training data acts as the control group in our study in order to explore the discriminability of learned features between clean test data and adversarial test data.
From the first row of Figure~\ref{fig:t-SNE}, which are the feature distributions of original training and test data of ‘feature(33)\_relu’ layer, ‘feature(43)\_relu’ layer, and ‘attention module’ layer, respectively. we can observe that with the increasing of network layer and the using of the attention mechanism, the feature distributions of the original test data and the training data are gradually approaching, and the accuracy of the model also rises, for example, from the 77.6\% to 85.2\%, which indicates the model learns the characteristic of the data well and classify it. However, from the feature distributions of adversarial test data in second row we can identity, with the increasing of network layer, their feature distributions are completely opposite to the original testing set, indicating that all features are classified incorrectly. As the dimension rises, the pre-trained classifier has a worse discriminative effectiveness on the adversarial image, which indicates that the perturbation information is more pronounced in the higher dimension than in the low dimension.

\textbf{Discussion:} In deep diagnostic models, it is a common practice to use more network layers to achieve better prediction performance. However, according to the analyses in Component 3, the adversarial perturbation information is more pronounced in higher dimensions. This reminds us that adding more network layers may not help. Besides, it is necessary to perform robust defense to avoid perturbation representations in high dimensions, and use adversarial training strategies to learn these perturbed representations.

\subsection{Multi-label Classification}\label{multi-label}
In the above three components, we present the performance of two models ({\em i.e.}, IPMI2019-AttnMel and Inception\_v3) under adversarial attacks for single-label classification. 
Now we investigate the performance of CheXNet under adversarial perturbations in a more challenging task, {\em i.e.}, multi-label classification (with each image annotated by multiple labels). 

%%%%%%%%%%%%%-------  New Section -------%%%%%%%%%%%%%
\noindent\textbf{Component 4: Label Correlation Analysis}

Proposing methods for evaluating the robustness of multi-label classifiers is a rarely touched and challenging task~\citep{song2018multi}. 
In our study, we attempt to evaluate the robustness of deep diagnostic model in multi-label classification from two aspects: (1) showing the quantitative classification results achieved by a deep network under adversarial perturbations; (2) visualizing the correlation of labels estimate for both original and adversarial images.

(1) \textbf{Quantitative Results}.
We apply both PGD attack and GAP attack to the CheXNet model, and report the results of multi-label classification in Table~\ref{table:multi-label adversarial_result}. This table shows that both ACC and AUC values yielded by CheXNet decrease greatly when faced with PGD and GAP attacks, indicating that CheXNet is not robust to adversarial perturbations.
To better illustrate the impact of adversarial perturbations on the multi-label classification problem, we show a chest X-ray image in Figure~\ref{fig:multi-label attack result}(a). The labels of this image are ``Effusion'', ``Cardiomegaly'', and ``Atelectasis'', with the probabilities of $95.02\%$, $97.16\%$, $72.56\%$, respectively. With both ``Hernia'' and ``Mass'' as the target labels, we use the PGD attack to produce adversarial image shown in Figure~\ref{fig:multi-label attack result}(c), while the perturbation is shown in Figure~\ref{fig:multi-label attack result}(d). We can see that even a slightly perturbed image can cause the multi-label CheXNet classifier to output wrong labels with high probabilities ({\em e.g.}, $92.33\%$ for ``Hernia'' and $97.51\%$ for ``Mass''). Besides, the probabilities of original true labels drop sharply to $0.02\%$ for ``Effusion'', $0.52\%$ for ``Cardiomegaly'', and $2.29\%$ for ``Atelectasis''. 
This reveals that despite the excellent diagnostic performance, the multi-label classification model CheXNet has low robust performance under adversarial attacks.
\begin{table}[!tbp]
	\setlength{\abovecaptionskip}{5pt}
	\setlength{\belowcaptionskip}{10pt}
	%	\footnotesize
	\centering
	\caption{Numerical results on ChestX-Ray dataset by PGD attack and GAP attack, respectively. The clearn ACC, AUC, and FR represent the input of original images. ACC and AUC are decreasing sharply under attacks while FR is increasing.}
	
	\centering  
	\resizebox{0.8\textwidth}{0.1\textheight}{
		\begin{tabular}{c c cc  c cccc}
			
			\toprule
			& \multirow{3}{*}{Clean}  & \multicolumn{2}{c} {\textbf{PGD Attack}}&  &\multicolumn{4}{c} {\textbf{GAP Attack}}  \\  % ?????
			\cmidrule{3-4} \cmidrule{6-9}
			
			&& $\epsilon = 2.0/255$ & $\epsilon=4.0/255$ &  & \multicolumn{2}{c}{$L_\infty = 7$} & \multicolumn{2}{c}{$L_\infty = 11$}
			\\
			&& $t = 1.0$ & $t = 4.0$ &  & X &val. & X& val.
			\\
			
			\midrule
			\specialrule{0em}{4pt}{4pt}
			%			\multirow{3}{*}{\begin{sideways}\textbf{ChestX-Ray}\end{sideways}}
			%			\specialrule{0em}{10pt}{10pt}
			ACC. & 86.5\% &64.1\% & 45.5\%&&56.4\%&58.0\%&44.6\%&39.5\% \\
			\specialrule{0em}{3pt}{3pt}
			AUC. & 0.807 &0.562& 0.308&&0.798&0.736&0.729&0.748 \\
			\specialrule{0em}{3pt}{3pt}
			FR. & - &23.6\%& 43.2\%&&29.7\%&33.1\% &42.7\%&51.4\% \\
			
			\specialrule{0em}{3pt}{3pt}
			
			\bottomrule
		\end{tabular}
		\label{table:multi-label adversarial_result}
	}
\end{table}

\begin{figure}[t]
	\setlength{\abovecaptionskip}{0pt}
	\setlength{\belowcaptionskip}{0pt}
	\setlength{\abovedisplayskip}{0pt}
	\setlength{\belowdisplayskip}{0pt}
	\centering
	\includegraphics[width=\textwidth]{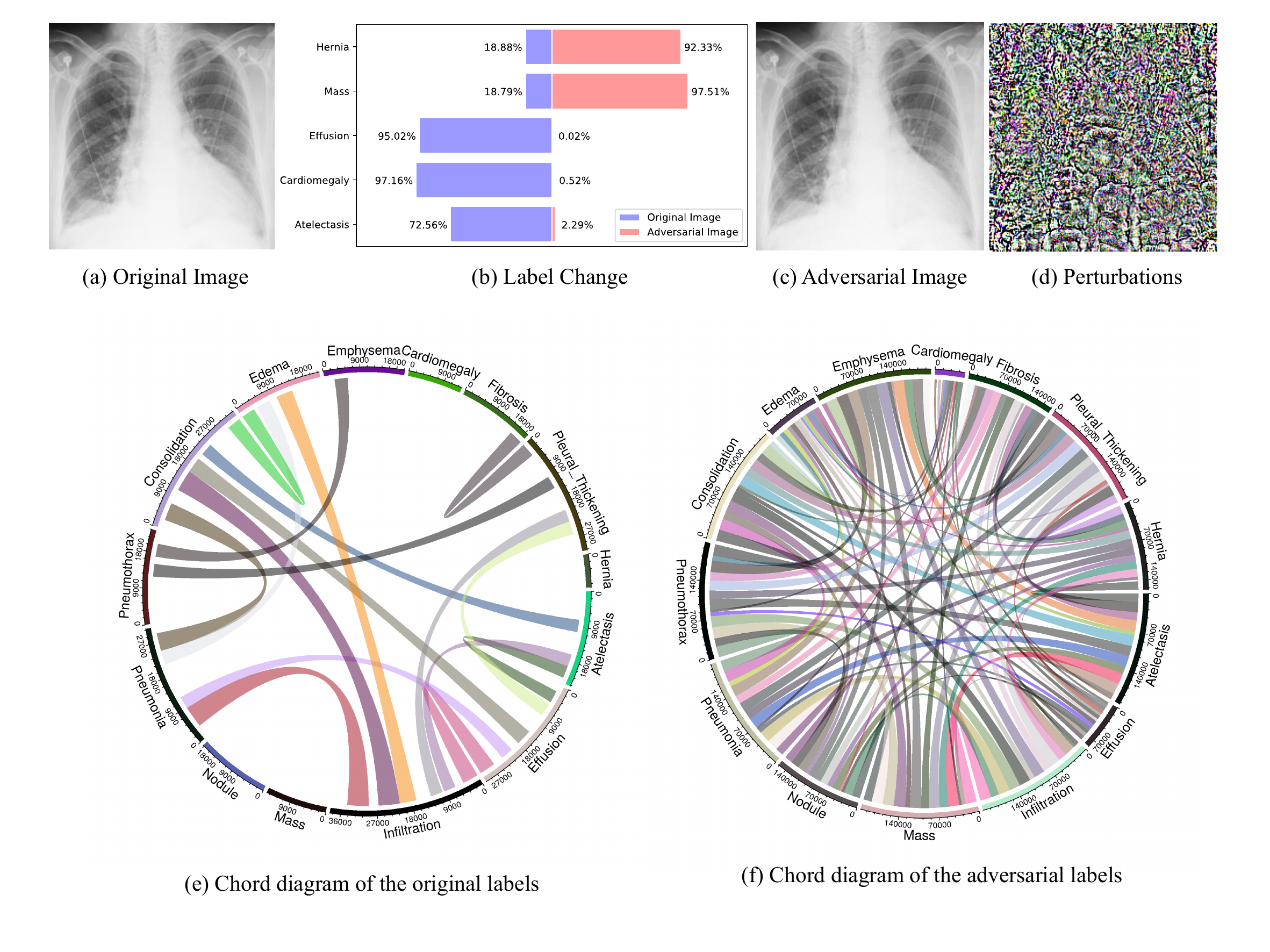}
	\caption{Correlation of  labels estimated by CheXNet for the original image and its adversarial image under PGD attack. (a) The original image as the input of CheXNet. (b) Labels estimated for the original and adversarial images. (c) The adversarial image attacked by PGD. (d) The small perturbation added to the original image. (e) The chord diagram of labels for the original images in the ChestX-ray14 dataset. (d) The chord diagram of labels for the adversarial images.}
	\label{fig:multi-label attack result}
\end{figure}

(2) \textbf{Visualization of Label Correlation}. In order to show our results more clearly, we use the chord diagram to compare the correlation of labels estimated for the original images and the adversarial images.
A chord diagram is a graphical method of displaying the inter-relationships between entities in a matrix. The data is arranged radially around a circle with the relationships between the data points typically drawn as arcs connecting the data. Here, such a diagram is based on the co-occurrence matrix of estimated labels, with each element in the matrix denoting the frequency of two labels simultaneously appears in an image.
We calculate the co-occurrence matrix on $22,433$ test images in the ChestX-ray14 dataset. In order to show our results more clearly, we use  elements with values greater than $3,000$ in the co-occurrence matrix.
We show the chord diagrams of estimated labels for the original and adversarial images in Figure~\ref{fig:multi-label attack result}(e)-(f), respectively. From Figure~\ref{fig:multi-label attack result}(e)-(f), one can observe that the number of occurrences of disease labels has changed dramatically. For instance, the original label of ``Infiltration'' appears $38,000$ times while the adversarial one appears $210,000$ times. Also, the correlation between labels is more tight for adversarial images, compared with that for original images. The possible reason is that, during the attacking process, the classifier's loss function is encouraged to become smaller toward irrelevant labels in order to make the classifier output wrong results. 
The most obvious finding from the analysis is that even if it is difficult to coordinate multiple labels, adversarial perturbations can easily change the relationship between labels and even establish new connections that never exist between labels.

\textbf{Discussion:} The label correlation of multi-label problems becomes closer after attack, which may make the model more likely to output wrong labels (because the probability of each label is similar). To this end, we can use a specially designed training strategy, that is, adding regularization terms to assign different weights to different labels. In this way, the model will pay more attention to those more important (e.g., with high weights) labels to improve robustness.

\subsection{Benchmark Results}\label{sec:benchmark Results}
The FP values of three deep diagnostic models on the \emph{Robust-Benchmark} dataset are reported in Table~\ref{table: fp}, respectively. As can be seen from Table~\ref{table: fp}, on perturbed inputs, three models are not robust. For example, the CheXNet on the Scale perturbation sequences have a $38.09\%$ probability of flipping between adjacent frames ({\em i.e.}, $FP^{CheXNet}=38.09\%$). 
\begin{table}[tbp]
	\setlength{\abovecaptionskip}{5pt}
	\setlength{\belowcaptionskip}{10pt}
	\centering
	\footnotesize
	\caption{%
		Flip probability (FP) of three deep diagnostic models on the \emph{Robust-Benchmark} dataset, respectively.%
	}
	\centering  
	%	\resizebox{0.49\textwidth}{0.03\textheight}{
	\begin{tabular}{cccc}
		
		\toprule
		Model & IPMI2019-AttnMel & Inception\_v3 & CheXNet \\
		\midrule
		FP & 9.52\% & 12.36\% & 38.09\%  \\
		\bottomrule
	\end{tabular}
	%	}
	\label{table: fp}
\end{table}

\begin{table}
	\centering
	\setlength{\abovecaptionskip}{5pt}
	\setlength{\belowcaptionskip}{10pt}
	\caption{The Relative Flip Probability (RFP) of models on \textit{melanoma Robust-Benchmark} dataset. Taking IPMI2019-AttnMel model as the benchmark model, the higher the RFP value, the lower perturbation robustness of the model. Ori\_ACC: accuracy with original test dataset; Adv\_ACC: accuracy with adversarial dataset by $\epsilon = 4.0/255, t=4.0$ PGD attack; RFP: relative flip probability with \textit{robust-benchmark} dataset.}
	\begin{tabular}{cccc}
		\toprule
		& Ori\_ACC & Adv\_ACC & RFP\\
		\midrule
		IPMI2019-AttnMel & 87.50\% & 0.00\% & 1.00 \\
		VGG19 & 83.38\% & 52.79\% & 0.03 \\
		ResNet50 & 83.64\% &36.15\% & 0.12 \\
		ResNet101 & 84.43\% & 25.33\% & 0.21\\
		ResNet152 & 85.49\% & 15.30\% & 0.49 \\
		Inception\_v3 & 82.32\% & 15.83\% & 0.69\\
		\bottomrule
	\end{tabular}
	\label{tab:RFP}
\end{table}

To validate the effectiveness of FP and our proposed \textit{robust-benchmark} datasets, we calculate the Relative Flip Probability (RFP) values of VGG19, ResNet50, ResNet101, ResNet152, and Inception\_v3 of \textit{melanoma robust-benchmark} dataset in Table~\ref{tab:RFP}. 
The Relative Flip Probability (RFP) value represents the FP value of other models compared with the benchmark model (\ie, IPMI2019-AttnMel used in this experiment). 
The RFP of VGG19 can be calculated as $RFP^{VGG19}=FP^{VGG19}/FP^{IPMI2019-AttnMel}$.
The higher the RFP value, the lower perturbation robustness of the model.
From Table 4, we can have the following observations. First, the original accuracy of model on clean dataset increases as the network gets larger (\ie, Ori\_ACC=83.64\% of ResNet50 and Ori\_ACC=85.49\% of ResNet152), while the accuracy on adversarial dataset decreases as the network gets larger (\ie, Adv\_ACC=36.15\% of ResNet50 and Adv\_ACC=15.30\% of ResNet152). It implies that a relatively large (\eg, with more layers and network parameters) network would have less robustness against adversarial attacks. Besides, it can be seen from Table 4 that the results of RFP have a similar trend. For example, the RFP of ResNet50 is 0.12 with the original accuracy of 83.64\%, while the RFP of ResNet152 is 0.49 with the original accuracy of 85.49\%. These results suggest that as the network complexity and the original accuracy results increase, the common perturbation robustness of the deep diagnosis model will become worse. This is consistent with results of adversarial robustness, while the accuracy-robustness trade-off has been proved to exist in predictive models when training robust models~\citep{tsipras2018robustness,zhang2019theoretically}. Furthermore, the RFP of the IPMI2019-AttnMel model is much higher than that of other models. This reminds us that when constructing a medical diagnostic model, we should not blindly increase the layers of a network or add assistance modules, since they may lead to a decrease in the robustness of the model. Through the above analyses, we hope our \textit{robust-benchmark} datasets can serve as the benchmark to evaluate the common perturbation robustness of deep diagnostic models in a standard manner.

\subsection{Robustness After Defense}\label{sec:defense results}
We evaluate the robustness of all three  deep diagnostic models trained with defense methods against PGD attack. The accuracy of three models are reported in Table~\ref{table: defense}, where ``None'' denotes the accuracy on natural models without attack or defense, ``Attack'' denotes natural models with PGD attack (4-step PGD with $\epsilon =4.0/255$), and ``Standard'' denotes conventional adversarial training~\citep{madry2017towards}. From Table~\ref{table: defense}, one can observe that the classification results of three attacked models have been significantly improved after using the MPAdvT and MAAdvT. For example, for binary-class melanoma classification task, the defense accuracy of MPAdvT is $82.4\%$ while standard adversarial training is $80.2\%$ which is much better than the result ({\em i.e.}, $0.0\%$) obtained when the model receives the adversarial attack, and even is comparable with that without any attack. We can get the same conclusion in multi-label classifier CheXNet for $83.9\%$ of MPAdvT while $81.6\%$ of standard adversarial training. We also evaluate the effectiveness of our proposed MAAdvT in Inception\_v3, for the defense accuracy of MAAdvT is $34.0\%$ which is higher than MPAdvT ($31.1\%$) and Standard ($25.0\%$).
We also use the \textit{Robust-Benchmark} to evaluate the robustness of these models after defense training, the flip probabilities of IPMI2019-AttnMel, Inception\_v3, and CheXNet with MPAdvT method are 0.0\%, 0.0\% and 0.3\%, respectively, which dramatically decrease compared with the original ones.
These results demonstrate the robustness of deep diagnostic models can be improved when trained by defense method. Besides, we can observe that our proposed MPAdvT and MAAdvT are more effective than standard adversarial training.

\begin{table}[t]
	\setlength{\abovecaptionskip}{5pt}
	\setlength{\belowcaptionskip}{10pt}
	\centering
	\footnotesize
	\caption{
		Defense accuracy of three deep diagnostic models on three datasets. ``None'' represents natural models without any attack or defense, ``Attack'' means natural models with PGD attack, ``Standard'' is natural models with standard adversarial training~\citep{madry2017towards} which has fixed perturbation and iteration step during training process. ``Attack'', ``Standard'', ``MPAdvT'' and``MAAdvT'' are tested with adversarial images while ``None'' tested with original images. Note that MAAdvT is only for single-label classification.
	}
	\label{table: defense}
	\centering  
	%	\resizebox{0.49\textwidth}{0.03\textheight}{
	\begin{tabular}{cccc}
		\toprule
		Model & IPMI2019-AttnMel & Inception\_v3 & CheXNet \\
		\midrule
		None & 87.5\% & 89.0\% & 86.5\%  \\
		Attack & 0.0\% & 0.0\% & 45.0\% \\
		Standard & 80.2\% &25.0\% & 81.6\% \\
		\textbf{MPAdvT} & \textbf{82.4\%}&\textbf{31.1\%} & \textbf{83.9\%} \\
		\textbf{MAAdvT} & \textbf{82.8\% }&\textbf{34.0\%} & \textbf{-}\\
		\bottomrule
	\end{tabular}
	\centering
	%	}
\end{table}

%% %% %% ---------------------------------------------------%% %% ^_^
\section{Discussion}
\label{S6}
In this section, we first summarize the performance of three representative deep diagnostic models under adversarial attacks. Considering the importance of medical safety, we also analyze whether the robustness of deep diagnostic models can be improved by using defense methods. %, which will guide our future work.

\subsection{Model Performance under Adversarial Attacks}
In order to explore whether deep diagnostic models are still reliable under adversarial perturbations, we present four components to show their performance in three types of tasks ({\em i.e.}, binary, multi-class and multi-label classification) in the experiments. Specifically, the numerical results change greatly between original images and adversarial ones, which indicate these models are vulnerable to adversarial perturbations. We can also see that adversarial attacks not only change the network outputs, but also change the response area and extraction of features within the network. 
More terrible, as the data dimension increases, the effectiveness of adversarial perturbations is more strong, the response of the model to the error area is more obvious.
As for multi-label classifier, we also analyze the change of label correlation. Even if it is harder to attack than the single-label classifier, we can clearly see that the label correlation during the attacking process have changed dramatically.
When evaluating the common robustness of three deep diagnostic models by \emph{Robust-Benchmark} datasets respectively, we can find the FP scores of these models are extremely high, indicating that the robustness of these models is poor.

It is so terrible to find that three types of deep diagnostic models are all unstable under adversarial perturbations. This can lead to a huge disaster in clinical medical diagnosis.
For the proposed \emph{Robust-Benchmark} dataset, we hope that it can be the benchmark for subsequent efforts to improve the robustness of deep learning models for computer-aided disease diagnosis.

\subsection{Diagnosis Performance After Defense}
The robustness of deep models is closely related to medical safety which is essential in clinical practice. Therefore, avoiding or at least reducing the vulnerability of deep diagnostic models is highly desired. To this end, 
we design two defense methods (\ie, MPAdvT and MAAdvT) aiming to improve the robustness of deep diagnostic models on three datasets. Preliminary results on the IPMI2019-AttnMel, Inception\_v3 and CheXNet trained by our defense methods and standard adversarial training are shown in Table~\ref{table: defense}. 

From Table~\ref{table: defense}, one can observe  that our defense methods are more effective than standard one. Besides, the classification results of three attacked models have been significantly improved after using the defense method.
However, how to effectively improve the robustness of deep diagnostic model so that it can resist more powerful attacks is still a difficult problem. As shown in Table~\ref{table: defense}, even if we add the misclassification aware regularization (\ie, MAAdvT) to train the Inception\_v3, there is still a certain gap between the defense accuracy (\ie, $34.0 \%$) and the original accuracy (\ie, $89.0\%$).
Making network "provably" defend from perturbations is a new direction in adversarial robustness~\citep{cohen2019certified, salman2019provably, lecuyer2019certified}. In the future, we could investigate the effectiveness of differentiation of correctly classified/misclassified training examples in the recently proposed certified/provable robustness framework and explore the potential improvements brought by the differentiation of training examples. 

\section{Conclusion}
\label{S7}
In this work, we evaluated the robustness of deep diagnostic models by adversarial attack.
Specifically, we have performed two types of adversarial attacks to three deep diagnostic models in both single-label and multi-label classification tasks, and found that these models are not reliable when attacked by adversarial example.
We have further explored how adversarial examples attack the models, by analyzing their quantitative classification results, intermediate features, discriminability of features and correlation of estimated labels for original/clean images and those adversarial ones.

To evaluate robustness of deep diagnostic models in a standard way, we created a new dataset called Robust-Benchmark and calculated flip probability of all these models, we hope that it can be the benchmark for subsequent efforts to improve the robustness of deep learning models for computer-aided disease diagnosis. We have also shown through experiments that the use of our proposed defense methods (\ie, MPAdvT and MAAdvT) can significantly improve the robustness of deep diagnostic models against adversarial attacks, which will guide our future work to explore more robust diagnostic models.

%% %% %% ---------------------------------------------------%% %% ^_^
\section*{Acknowledgements}
M. Xu, T. Zhang, Z. Li and D.Zhang were supported by the National Key Research and Development Program of China (No.~2018YFC2001600, 2018YFC2001602, 2018ZX10201002), the National Natural Science Foundation of China (Nos.~61876082, 61861130366, 61732006 and 61902183), and the Royal Society-Academy of Medical Sciences Newton Advanced Fellowship (No.~NAF$\backslash$R1$\backslash$180371).

\footnotesize
\bibliography{refer}
\bibliographystyle{elsarticle-harv}

\end{document}